\documentclass{article}

\usepackage{microtype}
\usepackage{graphicx}

\usepackage{hyperref}

\usepackage{longtable}
\usepackage{array}
\usepackage[dvipsnames]{xcolor}

\usepackage[accepted]{icml2025}

\usepackage{fontawesome5}
\usepackage{hyperref}

\makeatletter
\newcommand{\github}[1]{%
   \href{#1}{\faGithubSquare}%
}
\makeatother

\usepackage[utf8]{inputenc} 
\usepackage[T1]{fontenc}    
\usepackage{hyperref}       
\usepackage{url}            
\usepackage{booktabs}       
\usepackage{amsfonts}       
\usepackage{nicefrac}       
\usepackage{microtype}      
\usepackage{xcolor}         
\usepackage[normalem]{ulem}
\useunder{\uline}{\ul}{}

\usepackage{graphicx}
\usepackage{amsmath}
\usepackage{amssymb}
\usepackage{caption}
\usepackage{subcaption}
\usepackage{amsthm}
\usepackage{wrapfig}
\usepackage[capitalize,nameinlink]{cleveref}
\usepackage{xspace}
\usepackage{pifont}

\newcommand{\xmark}{\ding{55}}

\def\ie{{\em i.e.,}\xspace}

\DeclareMathOperator*{\argmax}{arg\,max}

\theoremstyle{plain}
\newtheorem{theorem}{Theorem}[section]

\theoremstyle{definition}

\newtheorem{observation}[theorem]{Observation}
\theoremstyle{remark}

\newcommand{\appropto}{\mathrel{\vcenter{
  \offinterlineskip\halign{\hfil$##$\cr
    \propto\cr\noalign{\kern2pt}\sim\cr\noalign{\kern-2pt}}}}}

\newcommand{\kvcache}{KV Cache\xspace}
\newcommand{\qfilters}{Q-Filters\xspace}

\icmltitlerunning{Q-Filters: Leveraging QK Geometry for Efficient KV Cache Compression}

\begin{document}

\twocolumn[

\icmltitle{Q-Filters: Leveraging Query-Key Geometry \\ for Efficient Key-Value Cache Compression}

\icmlsetsymbol{equal}{*}

\begin{icmlauthorlist}
\icmlauthor{Nathan Godey}{sorbonne,inria}
\icmlauthor{Alessio Devoto}{sapienza}
\icmlauthor{Yu Zhao}{uedi}
\icmlauthor{Simone Scardapane}{sapienza} \\
\icmlauthor{Pasquale Minervini}{uedi,miniml}
\icmlauthor{Éric de la Clergerie}{inria}
\icmlauthor{Benoît Sagot}{inria} \\
\begin{center}
  \noindent\faGithub\ \href{https://github.com/NathanGodey/qfilters}{github.com/NathanGodey/qfilters}  
\end{center}

\end{icmlauthorlist}

\icmlaffiliation{sorbonne}{Sorbonne Université, Paris, France}
\icmlaffiliation{inria}{Inria, Paris, France}
\icmlaffiliation{sapienza}{Sapienza University of Rome}
\icmlaffiliation{uedi}{University of Edinburgh}
\icmlaffiliation{miniml}{Miniml.AI}

\icmlcorrespondingauthor{Nathan Godey}{nathan.godey@inria.fr}

\icmlkeywords{Machine Learning, ICML}

\vskip 0.3in
]
\printAffiliationsAndNotice{}




\begin{abstract}
Autoregressive language models rely on a Key-Value (KV) Cache, which avoids re-computing past hidden states during generation, making it faster. 
As model sizes and context lengths grow, the \kvcache becomes a significant memory bottleneck, which calls for compression methods that limit its size during generation.
In this paper, we discover surprising properties of Query (Q) and Key (K) vectors that allow us to efficiently approximate attention scores without computing the attention maps.
We propose \qfilters, a training-free \kvcache compression method that filters out less crucial Key-Value pairs based on a single context-agnostic projection.
Contrarily to many alternatives, \qfilters is compatible with FlashAttention, as it does not require direct access to attention weights.
Experimental results in long-context settings demonstrate that \qfilters is competitive with attention-based compression methods such as SnapKV in retrieval tasks while consistently outperforming efficient compression schemes such as Streaming-LLM in generation setups.
Notably, \qfilters achieves a 99\% accuracy in the needle-in-a-haystack task with a $\times32$ compression level while reducing the generation perplexity drop by up to 65\% in text generation compared to Streaming-LLM.

\end{abstract}

\section{Introduction}
The performance of Large Language Models (LLMs) as autoregressive text-generation systems relies on the effectiveness of the Transformer architecture \citep{transformers}. 
Recently, long-context models such as Gemini-Pro-1.5~\citep{reid2024gemini}, Claude-3~\citep{anthropic2024claude}, GPT-4~\citep{achiam2023gpt}, and Llama3.1~\citep{llama3-head} have demonstrated the ability to process hundreds of thousands of tokens. However, processing such long sequences comes with significant challenges, as it may lead to higher decoding latency and memory saturation. 
As the context length grows, each inference step involves storing an increasingly large context from GPU memory in the form of the \kvcache, creating a memory bottleneck that hinders efficient inference \citep{yaofu-long-context-challenge}.
%
%
\begin{figure}[t]
    \centering
    \includegraphics[width=0.45\textwidth]{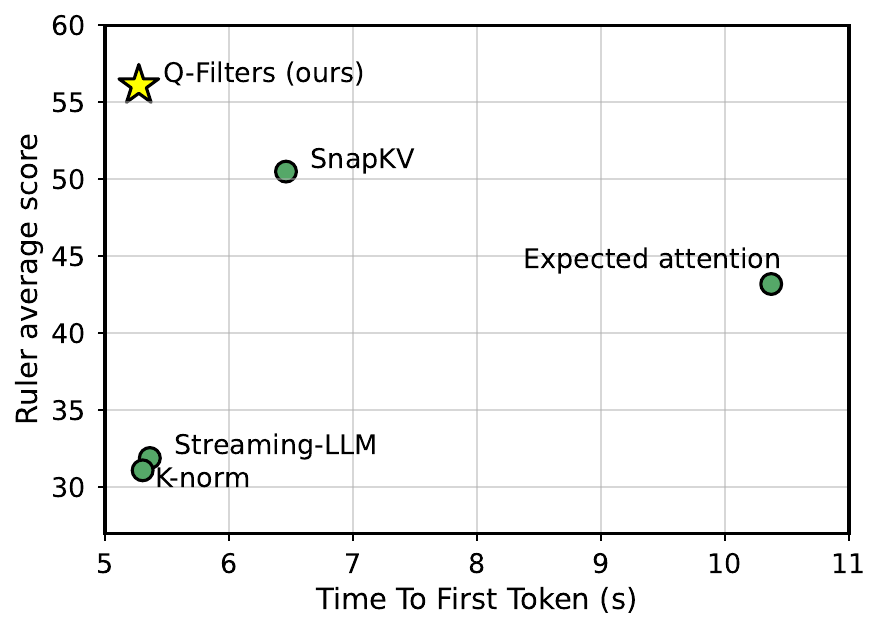}
    \caption{Accuracy vs Time to First Token (TTFT) tradeoff for Llama-3.1-70B-Instruct, measured on the Ruler dataset with $\times$8 compression. The TTFT is measured using 2 A100 GPUs on 8192-tokens sequences.}
    \label{fig:ttft}
\end{figure}
\begin{figure*}[ht]
    \centering
    \begin{subfigure}{0.33\textwidth}
         \includegraphics[width=\textwidth]{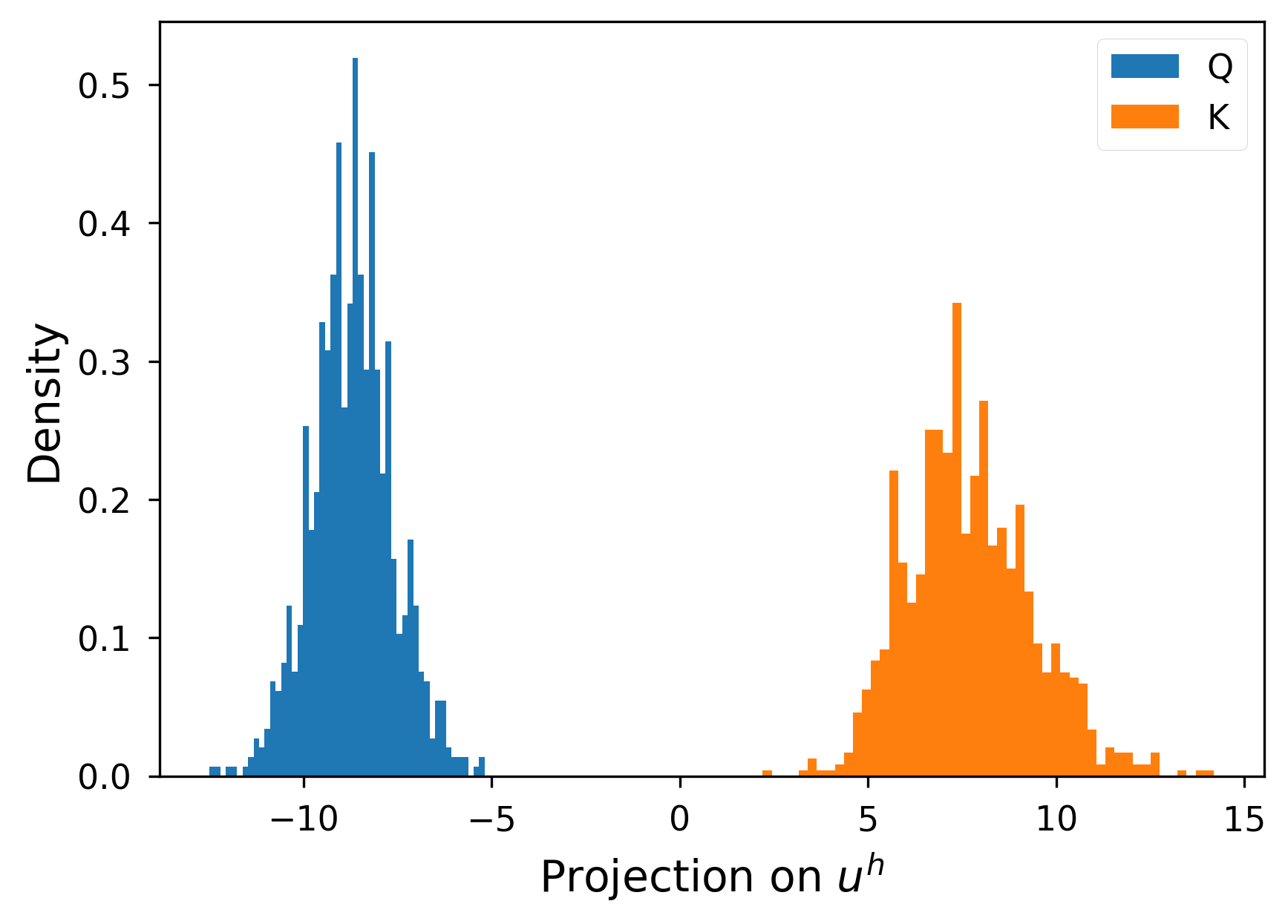}
         \caption{Layer 14, Head 5 ($\epsilon=-1$)}
         \label{fig:proj_l14_h5}
    \end{subfigure}
    \hfill
    \begin{subfigure}{0.33\textwidth}
         \includegraphics[width=\textwidth]{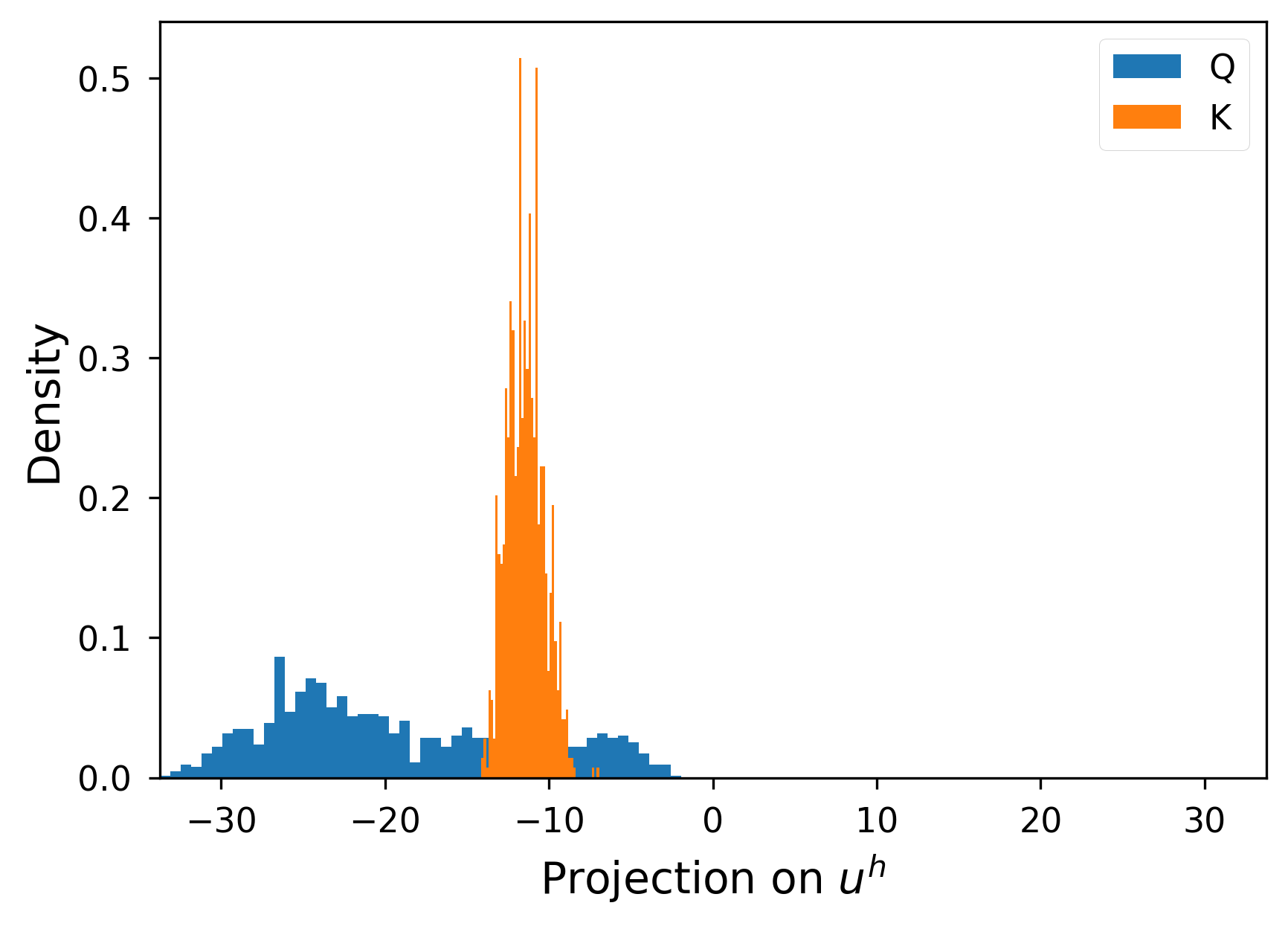}
         \caption{Layer 31, Head 14 ($\epsilon=+1$)}
         \label{fig:proj_l31_h14}
    \end{subfigure}
    \hfill
    \begin{subfigure}{0.33\textwidth}
     \centering
     \includegraphics[width=\textwidth]{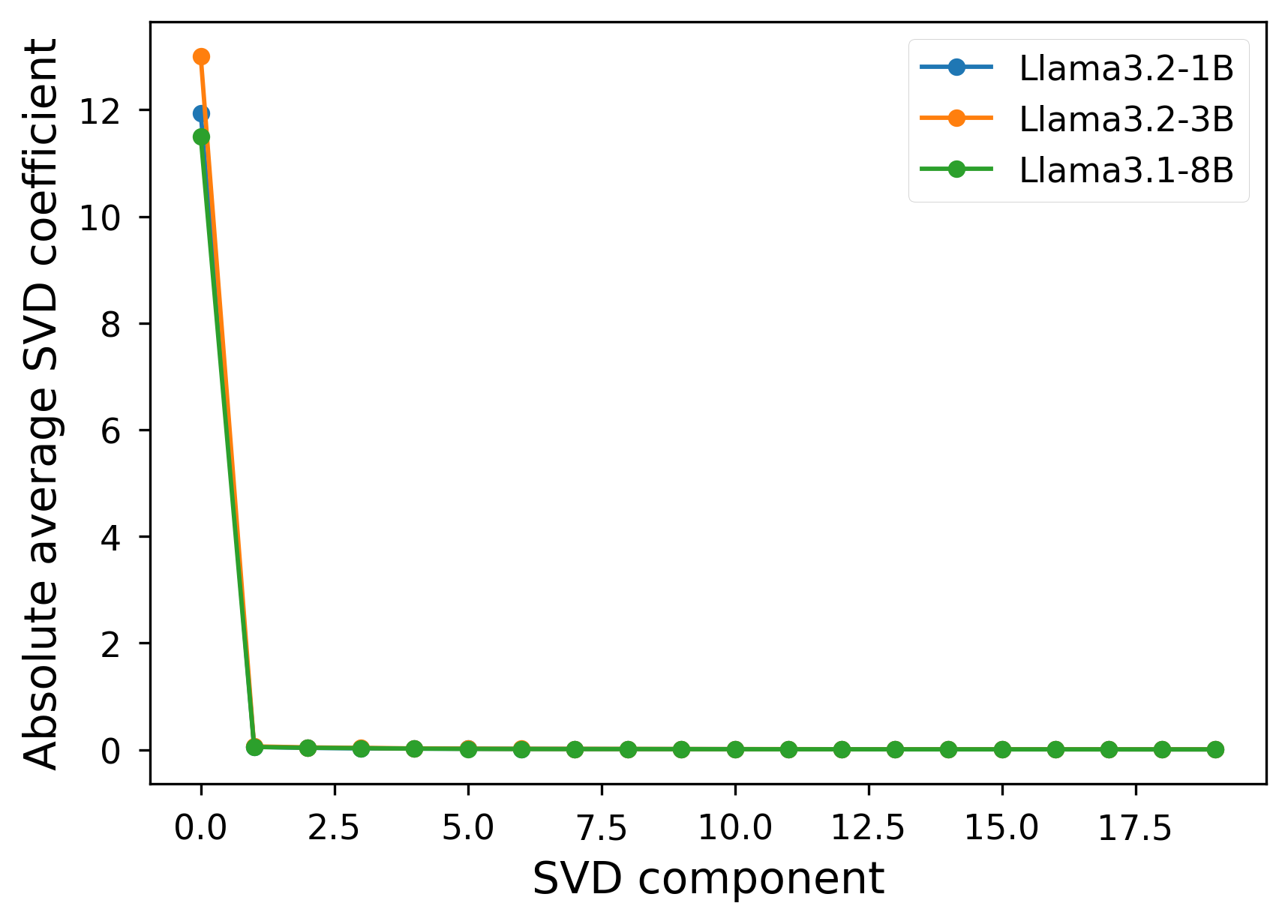}
     \caption{SVD absolute average coefficients}
     \label{fig:svd_coeff_avg}
    \end{subfigure}
    \caption{Left and center: distributions of the projections of $Q^h$ and $K^h$ on $u^h$ for Llama-3.1-8B. Right: estimates of $\left|\mathbb{E}_{i}(\langle Q^h_i, v_m \rangle)\right|$ where $v_m$ are the right vectors from the SVD of a set of $Q^h$ representations from different Llama models, averaged over all layers and heads. }
\end{figure*}
To address this issue, \kvcache compression methods aim to reduce the size of this past-context representations storage by removing or merging Key-Value pairs, thereby alleviating memory bottlenecks.
While \kvcache compression techniques have gained popularity, many approaches require fine-tuning or re-training the underlying models \citep{nawrot2024dynamic, ainslie2023gqa, deepseekaiv2}, which limits their applicability in real-world deployment scenarios.
%
%
Training-free methods have also been proposed, but they often rely on access to attention weights to evaluate the importance of Key-Value pairs \citep{xiao2024efficient, li2024snapkv}, making them incompatible with the widely adopted efficient attention algorithm FlashAttention~\citep{dao2023flashattention2}.
These methods usually require a partial re-computation of the attention matrices, which leads to a time and memory overhead. Hence, these algorithms are often used to compress prompts before generating answers and are not ideally suited for memory-constrained generation.
In this work, we propose \qfilters, a training-free \kvcache compression method that uses the geometric properties of \textbf{Q}uery-Key to \textbf{filter} out the less important Key-Value pairs. Our approach achieves competitive results across synthetic tasks and pure generation cases while maintaining compatibility with FlashAttention and, thus, better time efficiency.
%
%

Analysing the properties of queries ($Q$) and Keys ($K$) distributions, we find that \emph{a single direction, spanned by the principal eigenvector of $Q$, encodes an input selection process for each head}.
Identifying this direction allows us to efficiently estimate which inputs are mostly ignored by a given head and can thus be discarded with minimal performance loss.
Interestingly, we find that this direction is context-agnostic, \ie the directions we identify in different contexts are highly consistent.
Leveraging this property, we calculate lightweight projections, which we refer to as \qfilters, based on a small held-out calibration dataset only once for every model, incurring minimal computational overhead. 
%
%
At inference time, we use \qfilters to project Keys in the pre-computed direction to estimate the importance of  Key-Value pairs without accessing attention scores, and we prune the \kvcache accordingly.
This makes our method faster than most \kvcache compression alternatives that use attention scores to estimate the importance of the KV pairs.
%

Additionally, our method is training-free, requiring only a very short initial calibration, and we show it can be easily applied to a variety of decoder-only language models. 
%
We validate our method on a wide set of tasks, ranging from language modelling to in-context learning and long-context tasks, achieving competitive performance even with 32x compression ratios.
\begin{figure}[t]
     \centering
     \begin{subfigure}[h]{0.9\columnwidth}
         \centering
         \includegraphics[width=\textwidth]{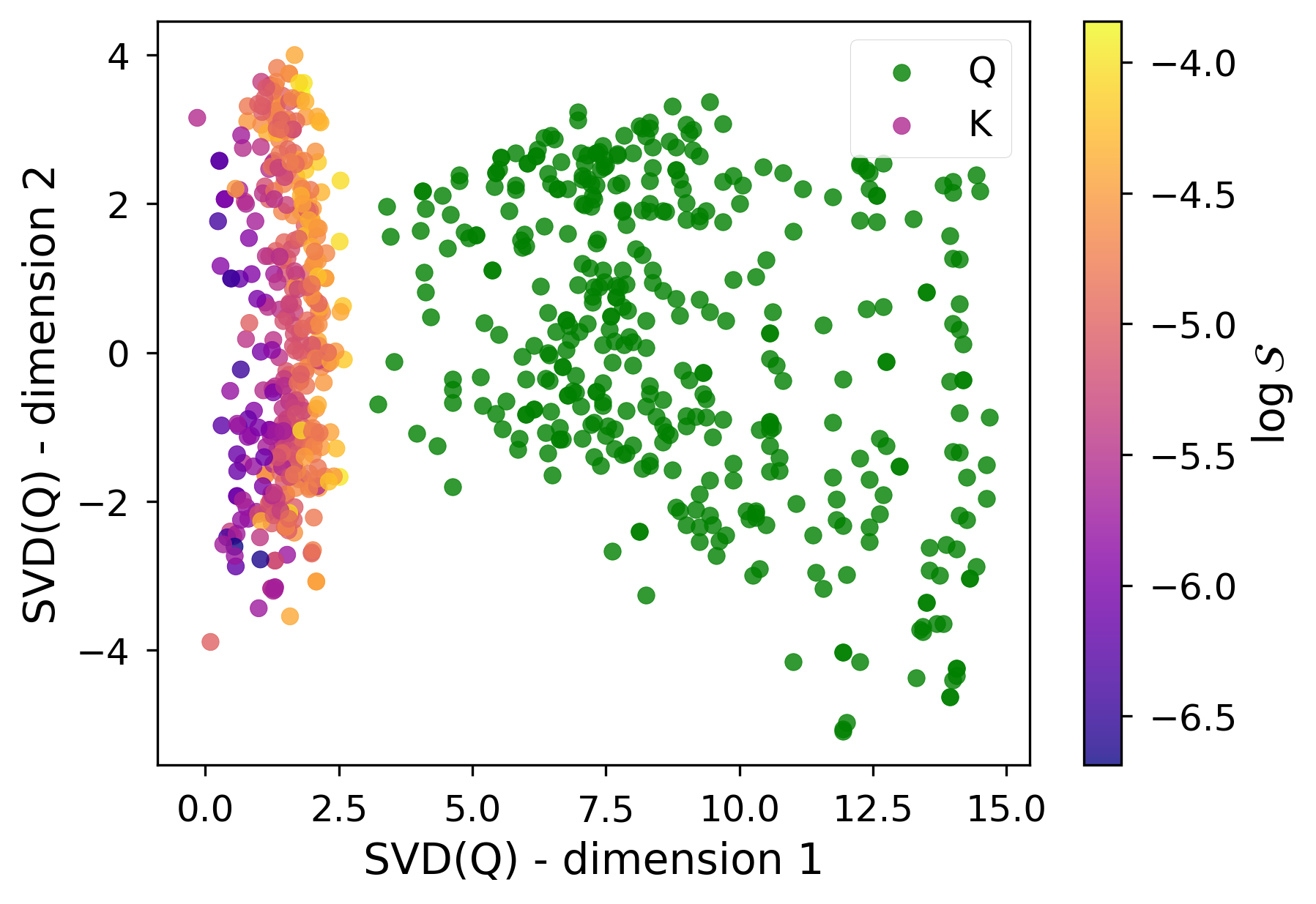}
         \caption{Layer 0, Head 21}
         \label{fig:ll1b_lay0_head21_svd}
     \end{subfigure}
     \begin{subfigure}[h]{0.9\columnwidth}
         \centering
         \includegraphics[width=\textwidth]{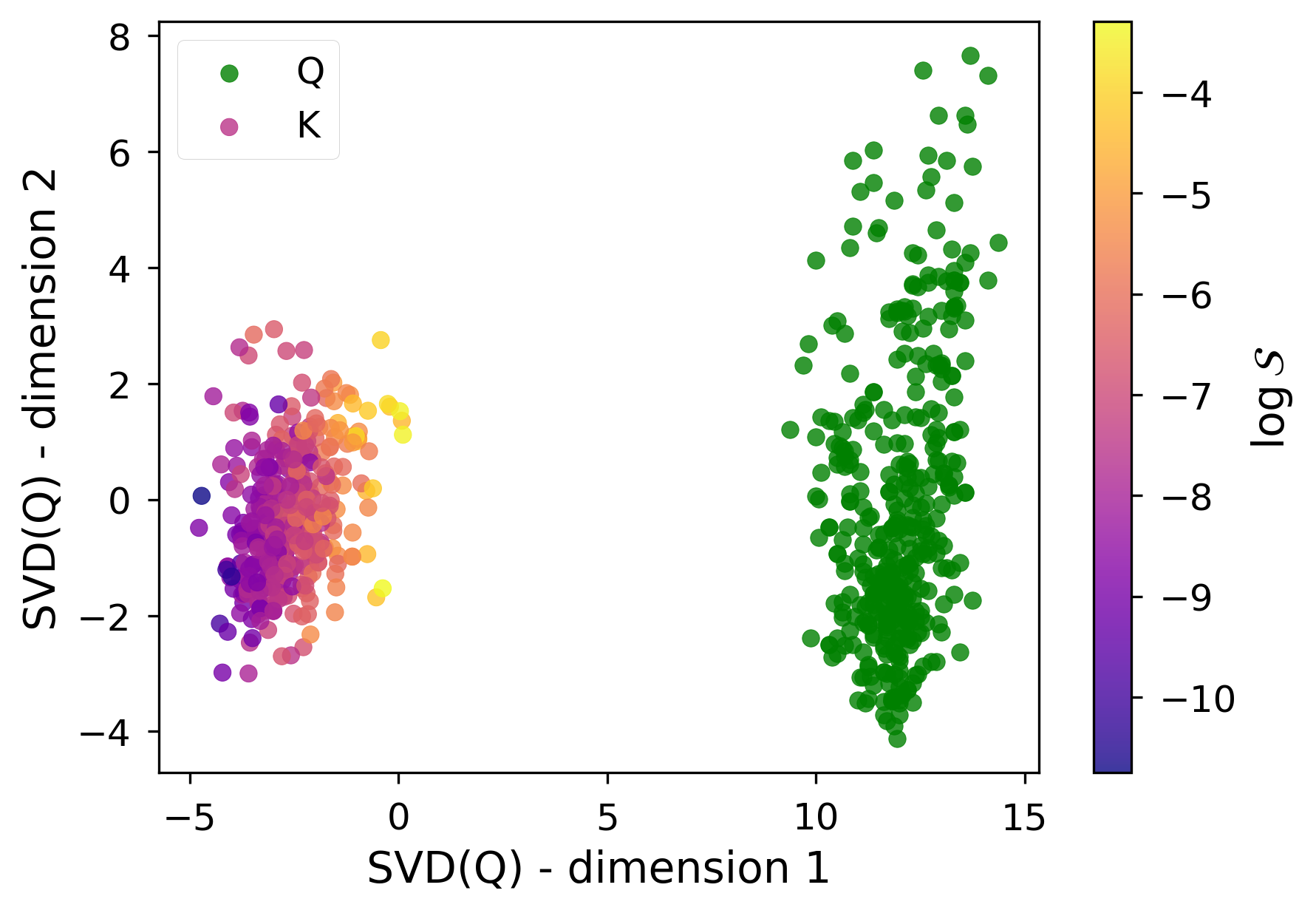}
         \caption{Layer 13, Head 16}
         \label{fig:ll1b_lay13_head16_svd}
     \end{subfigure}
        \caption{Projection of $Q^h$ and $K^h$ vectors in the first two components of the SVD of $Q^h$ for different heads in Llama-3.2-1B. The colour on the $K$ projections represents the $\log$-average attention at the corresponding index for the current head. The $x$-axis and $y$-axis indicate the results of a projection of the representations on $v_1$ and $v_2$, respectively.}
        \label{fig:ll1b_svd}
\end{figure}
\section{Background}
\subsection{Key-Value Cache}
 We first introduce the relevant notation for our analysis and the role of the \kvcache in efficient LLM inference. Consider a transformer model with a hidden dimension $d_m$ and $n_l$ layers, processing a sequence of length $L$. Each transformer layer processes the input sequence via Multi-Head Self-Attention (MHA).
In MHA, the model transforms the input features $X \in \mathbb{R}^{L \times d_m}$ into three distinct representations for each attention head $h \in [1, H]$. These representations, known as queries $Q^h$, Keys $K^h$, and Values $V^h$, each belong to $\mathbb{R}^{L \times d_h}$, where $d_H = d_m / H$ represents the dimension per head, and $h$ denotes the head index. The second step computes the attention output $O^h$ for each head using the following equation:
\[
O^h = \text{softmax} \left( \frac{Q^h (K^h)^T}{\sqrt{d_H}} \right) V^h.
\]
%
In causal modelling, where the model generates text sequentially, we ensure that each token only attends to previous tokens and itself. This causality constraint means that when generating the $t$-th token, its output $O^h_t$ depends only on the current and previous inputs, as expressed by:
\[
O^h_t = \text{softmax} \left ( \frac{Q^h_t (K^h_{\leq t})^T}{\sqrt{d_H}} \right) V^h_{\leq t}.
\]
The Key and Value representations $K^h_{\leq t}, V^h_{\leq t}$, which combine previous Keys and Values with the current ones $K^h_t, V^h_t$, reuse information from previous generation steps. By storing these representations in a \kvcache during the generation process, we can avoid the computational cost of recalculating them at each step, thereby significantly improving efficiency at the cost of the memory occupied by stored KV pairs.
However, this memory-compute tradeoff introduces a new challenge: as the context length grows, decoding latency increases due to the frequent transfers of large \kvcache states between high-bandwidth memory (HBM) and streaming multiprocessors (SM)~\citep{yaofu-long-context-challenge}. For this reason, \kvcache compression methods have become essential to allow inference in long contexts. 
%
%
%
%
%
\subsection{Geometry of Multi-Head Attention}

In \citet{devoto-etal-2024-simple}, the authors examined a relationship between basic characteristics of the Key representations and attention score distributions. Notably, they observe a negative correlation between the average attention weight given to a position and the $L_2$-norm of the $K^h_t$ vector at that position. 
Leveraging this observation, they propose to compress the \kvcache by selecting the KV pairs for which $||K^h_t||_2$ is the smallest. Using this simple heuristic, they are able to reach $\times 2$ compression ratios without altering the retrieval and modelling performance of the models they study.
In their paper, while they relate this approach to the well-known oulier dimension phenomenon \citep{kovaleva-etal-2021-bert}, they do not provide a grounded explanation as to the strength of the observed correlation.

A promising path towards a better explanation of the $L_2$-norm observation consists in systematically exploring the geometry of the representations involved in the attention score computation, namely $Q^h$ and $K^h$.

\citet{godey-etal-2024-anisotropy} show that the distributions of $Q^h_t$ and $K^h_t$ are \textit{anisotropic}, i.e. they do not uniformly occupy $\mathbb{R}^{d_H}$. They observe that both distributions ``drift away'' from the origin as training progresses. Crucially, this drift occurs along parallel directions in $\mathbb{R}^{d_H}$, so that the dot product between mean $Q^h_t$ and mean $K^h_t$ representations tends to increase in absolute value, and to be either positive or negative for different heads. In the paper, it is argued that this drift could be linked to the sparsity of attention patterns, but the authors do not propose a clear interpretation of this phenomenon from the perspective of the attention mechanism.

In this paper, we bridge the gap between the two aforementioned observations; namely, we explain the effectiveness of the $L_2$-norm heuristic introduced in \citet{devoto-etal-2024-simple} by leveraging the (jointly) anisotropic nature of Query-Key representations, and we explore a stronger heuristic that exploits this finding to refine the $L_2$-norm approximation by projecting Keys onto the drift directions, that we refer to as \qfilters.

\begin{figure}[t]
     \centering
     \begin{subfigure}[h]{0.8\columnwidth}
         \centering
         \includegraphics[width=\textwidth]{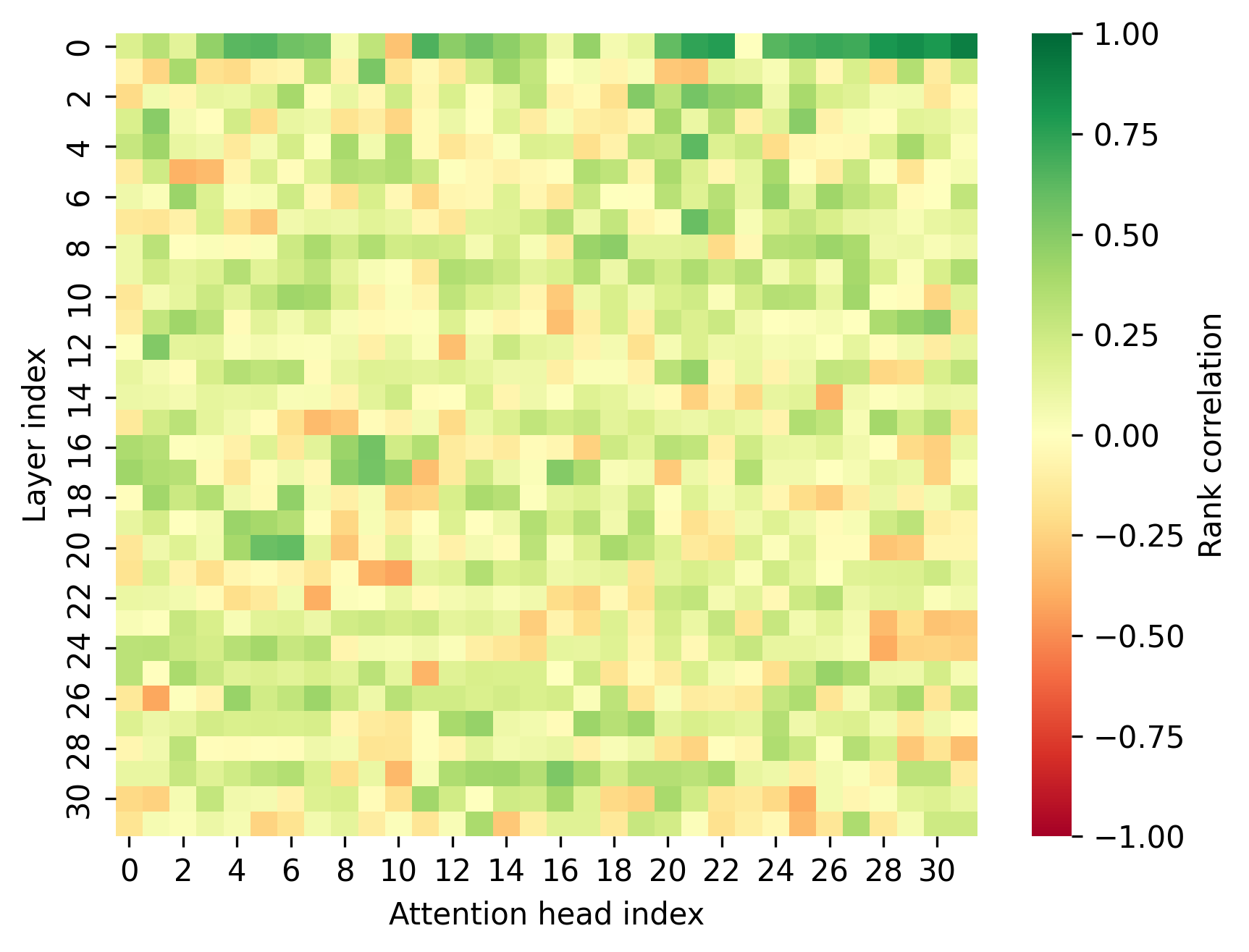}
         \label{fig:corr_info_norm}
     \end{subfigure}
     \begin{subfigure}[h]{0.8\columnwidth}
         \centering
         \includegraphics[width=\textwidth]{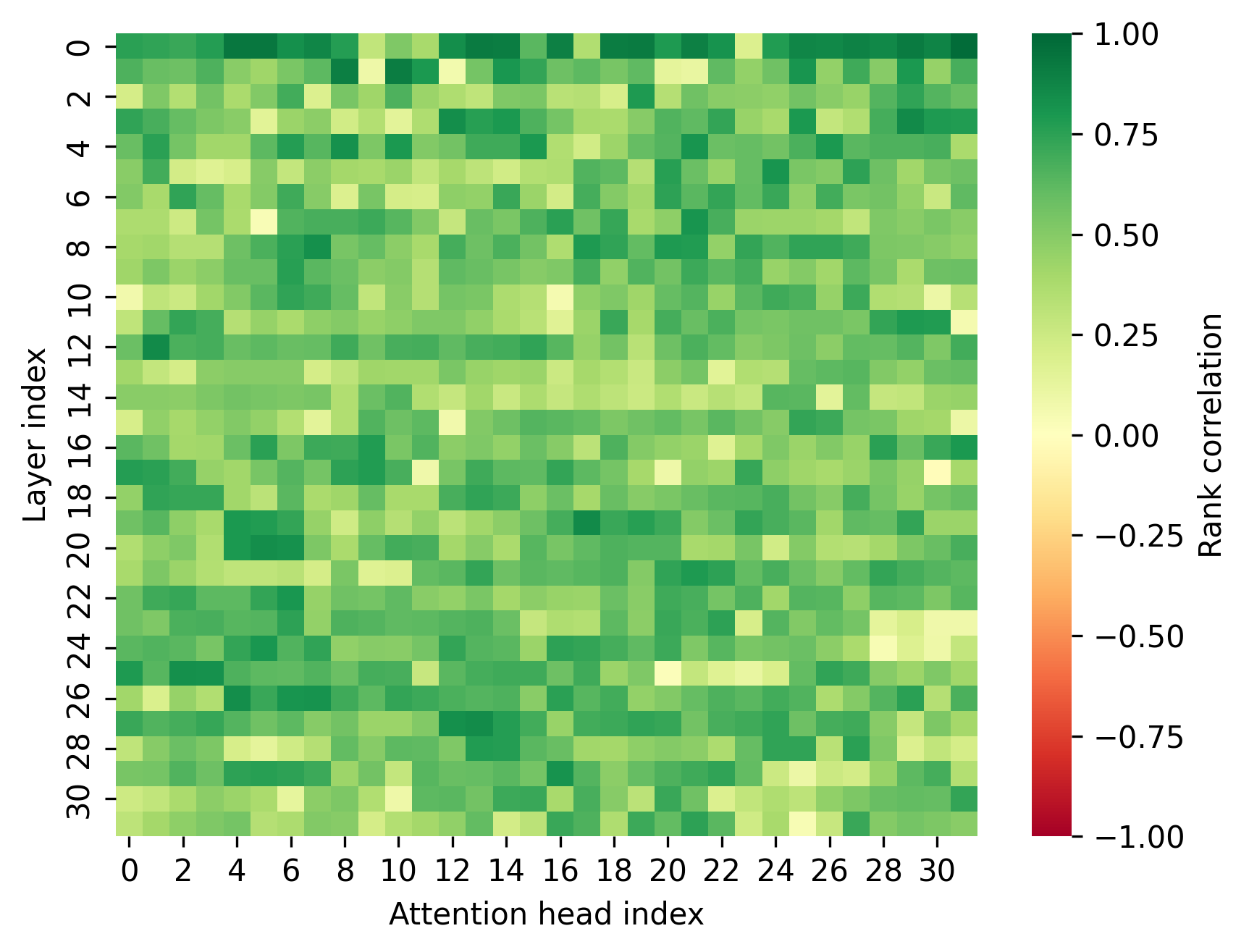}
         \label{fig:corr_info_svd}
     \end{subfigure}
        \caption{Spearman rank correlation between KV compression scoring metrics and the observed attention $S^h$ for Llama-3.2-1B, for K-norm (top) and \qfilters (bottom).}
        \label{fig:rank_corr}
\end{figure}
\section{Method}
\label{sec:method}
\subsection{Exploring the Query-Key Geometry}
Motivated by \citet{devoto-etal-2024-simple} and \citet{godey-etal-2024-anisotropy}, we propose to further explore some geometrical properties of $Q^h$ and $K^h$ vectors and their implications for unnormalized attention logits $Q^h (K^h)^T$.

First, we formalize the findings from \citet{godey-etal-2024-anisotropy} into our theoretical framework. The authors shed light on the existence of a \textit{favored} common normalized direction for both $Q^h$ and $K^h$ distributions. We denote such direction as $u^h \in \mathbb{S}^{d_H-1}$ where $\mathbb{S}^{d_H-1}$ is the $d_H$-dimensional hypersphere (i.e. $\mathbb{S}^{d_H-1} = \{ x \in \mathbb{R}^{d_H} \text{ s.t. } ||x||_2 = 1\}$). As a consequence, the projection of $Q^h$ and $K^h$ distributions on $u^h$ is usually non-null but can take opposite signs in $Q^h$ and $K^h$. Hence, we use $\epsilon=\pm 1$ to account for the possible sign discrepancy and formulate the following \cref{assum:1} in terms of expectation.
\begin{observation}[Joint anisotropy]
\label{assum:1}
There exist $u^h \in \mathbb{S}^{d_H - 1}$ and $\epsilon=\pm 1$ such that
\begin{equation*}
\mathbb{E}\left(\langle Q^h_i, u^h \rangle\right) > 0 \quad \text{and} \quad \mathbb{E}\left(\langle K^h_j, \epsilon u^h \rangle\right) > 0,
\end{equation*}
where $\langle \cdot, \cdot \rangle$ denotes the dot product.
%
\end{observation}
To validate \cref{assum:1}, we compute the Singular Value Decomposition (SVD) of a set of $Q^h$ representations taken from various sequences for Llama-3.1-8B. We find that the first right-vector of the SVD verifies \cref{assum:1} for all tested heads, and we display examples of projection distributions in \Cref{fig:proj_l14_h5,fig:proj_l31_h14}. The intuitive consequence of this observation regarding attention weights is that, if a given $K^h_t$ has a strong projection along $\epsilon u_h$, then future queries $Q^h_{\geq t}$ can be expected to have a stronger dot-product with $K^h_t$ in average.

However, it is not clear \textit{a priori} that this effect is uni-directional, i.e. that there exists a unique direction $u^h$ (up to a sign) that verifies \cref{assum:1}. Hence, identifying one such direction may not suffice to characterize the anisotropy of $Q^h$ representations and to derive estimations of the dot-products used in attention. The uni-directional nature of the Query-Key anisotropy can be formalized as in \cref{assum:2}.

\begin{observation}
\label{assum:2}
Let $u^h = \argmax_{u \in \mathbb{S}^{d_H - 1}} \mathbb{E}\left(\langle Q^h_i, u \rangle\right)$ and $B=(u^h, u_2, ..., u_{d_H})$ an orthonormal basis of $\mathbb{R}^{d_H}$. Then for all attention inputs $X$:
$$
\forall m \in [2, d_H],
    \mathbb{E}\left(\langle Q^h_i, u_m \rangle \right) \approx 0
$$
\end{observation}

In \Cref{fig:svd_coeff_avg}, we observe that only the first singular component of the SVD of $Q^h$ representations carries an anisotropic behavior, as the projections on all other components have a null mean. Hence, by taking the SVD right-vector basis as $B$, we can show that the first component of the SVD empirically verifies \cref{assum:2}.
%
%
This lets us derive a basic estimation for the average unnormalized attention logits $\langle Q^h_i, K^h_j \rangle$.
\begin{theorem}[{\normalfont proof in \Cref{app:proof}}]
\label{thm:main}
Under \Cref{assum:1} and \Cref{assum:2}, we have:
$$
\mathbb{E}_{Q^h_i}(\langle Q^h_i, K^h_j \rangle) \approx \kappa^h \langle K^h_j, u^h \rangle
$$
where $\kappa^h$ is a positive constant.
\end{theorem}
%
Intuitively, projecting $K^h_t$ along the anisotropic direction $u^h$ gives us an estimate of the attention logits that involve $K^h_t$ up to a positive multiplicative constant $\kappa^h$.

This result provides a justification for the method developed in \citet{devoto-etal-2024-simple}. As a matter of fact, \Cref{assum:1} implies that $\mathbb{E}_j\left(\cos(K^h_j, u^h)\right)$ should have the same sign as $\epsilon$. In practice, we observe $\epsilon=-1$ for a vast majority of heads in trained causal LMs. Hence, we can derive a looser estimation from \Cref{thm:main}:
$$
\mathbb{E}_{i, X}(\langle Q^h_i, K^h_j \rangle) \approx -\kappa^h \left|\mathbb{E}_{j, X}\left(\cos(K^h_j, u^h)\right)\right| ||K^h_j||_2
$$
This estimation shows that the $L_2$-norm of $K^h_j$ vectors is negatively correlated with the corresponding mean attention logits and can therefore be used to approximate them. However, only using the $L_2$-norm to estimate the attention score as done in \citet{devoto-etal-2024-simple} is suboptimal, as it ignores the angular component of the $\langle K^h_j, u^h \rangle$ product.
In practice, one can approximate $u^h$ as defined in \Cref{assum:2} using the SVD of concatenated representations $Q^h$ extracted by passing samples through the model.
Formally, we collect a batch of Query activations $\mathcal{Q}^h = \{Q^h_1, Q^h_2, ..., Q^h_n\}$ by passing documents sampled from pre-training corpora and using the right-vectors $V$ as the orthonormal basis $B$:
\begin{equation}
\label{eq:qfilters}
    \mathcal{Q}^h = U \Sigma V^\top, \text{with } V = (v_1, v_2, ..., v_{d_H})
\end{equation}
The resulting $v_1$ vectors are, up to a sign, what we refer to as \qfilters, as they allow to estimate which Key-Value pairs are worth storing for each head along generation.
%
%
%
%
%
%
%
%
%
%
%
%
\Cref{fig:ll1b_svd} also displays information about attention levels for the corresponding indices. For a given input $X$, we measure the average attention at position $t$ as:
$$
\mathcal{S}^h_t = \frac{1}{L-t+1} \sum_{i=t}^L A^h_{it},
$$
where $A^h$ is the attention map for head $h$. It appears clearly from \Cref{fig:ll1b_svd} that there exists a strong correlation between the average attention at a given index and the projection of $K^h$ on the $v_1$ component.

We observe that the projection of $K^h$ on the $v_1$ component has a consistent sign for a given head, e.g., it is consistently positive in \Cref{fig:ll1b_lay0_head21_svd} and consistently negative in \Cref{fig:ll1b_lay13_head16_svd}, while the projection results on $v_2$ have a near-zero expectation, further validating \Cref{assum:1} and \Cref{assum:2}.
\subsection{Q-Filters}
Based on \Cref{thm:main}, we can design a \kvcache compression scheme that consists of the following steps:
\begin{enumerate}
\item For a given model, retrieve its \qfilters, which can be obtained with the following procedure:
\begin{enumerate}
    \item Gather $Q^h$ representations by passing samples through the model;
    \item Compute the SVD of the gathered representations at each layer and head;
    \item Obtain the \textit{positive} right vector (or Q-Filter) for each head $v_1^+ = \text{sgn}(\mathbf{1} u_1^T) v_1 $.
\end{enumerate}
\item At inference, for each head, discard the $K^h_t$ with the lowest $\langle K^h_t, v_1^+ \rangle$ value.
\end{enumerate}
In the case of Grouped-Query Attention or GQA \citep{ainslie2023gqa}, we simply average the \qfilters for each group of Query representations.

We bring the attention of the reader to the fact that this method only requires a single preparation step following training for a given model. The \qfilters are entirely context-agnostic and rely on inherent properties of the Query and Key latent spaces. In the rest of this article, we use a subset of the Pile dataset \citep{pile} to compute the Q-Filters and discuss the choice of the dataset and of the number of necessary SVD samples in \Cref{sec:calib_dataset}.
In \Cref{fig:rank_corr}, we observe that the \qfilters heuristic is noticeably more correlated with the attention score $S^h$ for most heads compared to the $L_2$-norm metric. As such, ordering the Key-Value pairs using the \qfilters heuristic should allow us to select more relevant pairs than using the method from \citet{devoto-etal-2024-simple} - that we will call $K$-norm for the sake of simplicity.
\section{Experiments}
We validate our method both on memory-constrained language modelling and on long-context retrieval tasks (e.g. needle-in-a-haystack). Additionally, we test our method on the Ruler dataset \cite{hsieh2024ruler}, which is specifically designed to test the model's long context modelling abilities. 
We test \qfilters on Llama-3.1-8B, Llama-3.1-70B \cite{llama3-head} and Qwen-2.5-7B \cite{qwen2025qwen25technicalreport}, but the method can be easily adapted to any pre-trained decoder-only LLM.
We compare \qfilters with several \kvcache compression methods. These include StreamingLLM~\citep{xiao2024efficient}, which focuses on language modeling by always retaining the initial tokens of the sequence. We also compare with SnapKV~\citep{li2024snapkv}, which performs compression based on attention scores from the final portion of the prompt, making it particularly suitable for compression of large prompts. Additionally, we compare against preserving low-$L_2$ norm tokens~\citep{devoto-etal-2024-simple} and the recent ExpectedAttention~\citep{kvpress2024}.

\begin{figure}[t]
     \centering
     \begin{subfigure}[h]{0.85\columnwidth}
         \centering
         \includegraphics[width=\textwidth]{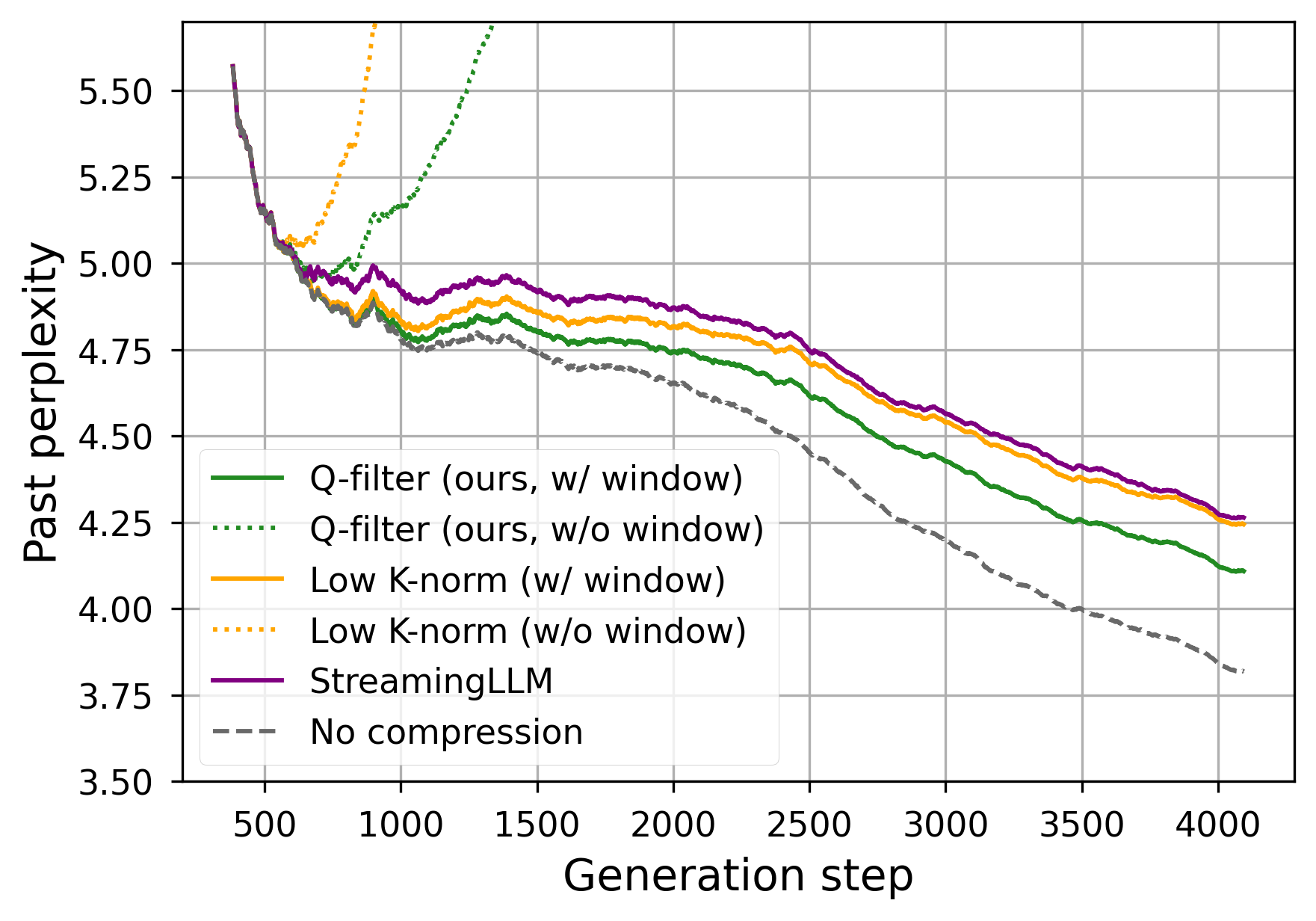}
         \label{fig:stream_8B}
     \end{subfigure}
     \begin{subfigure}[h]{0.85\columnwidth}
         \centering
         \includegraphics[width=\textwidth]{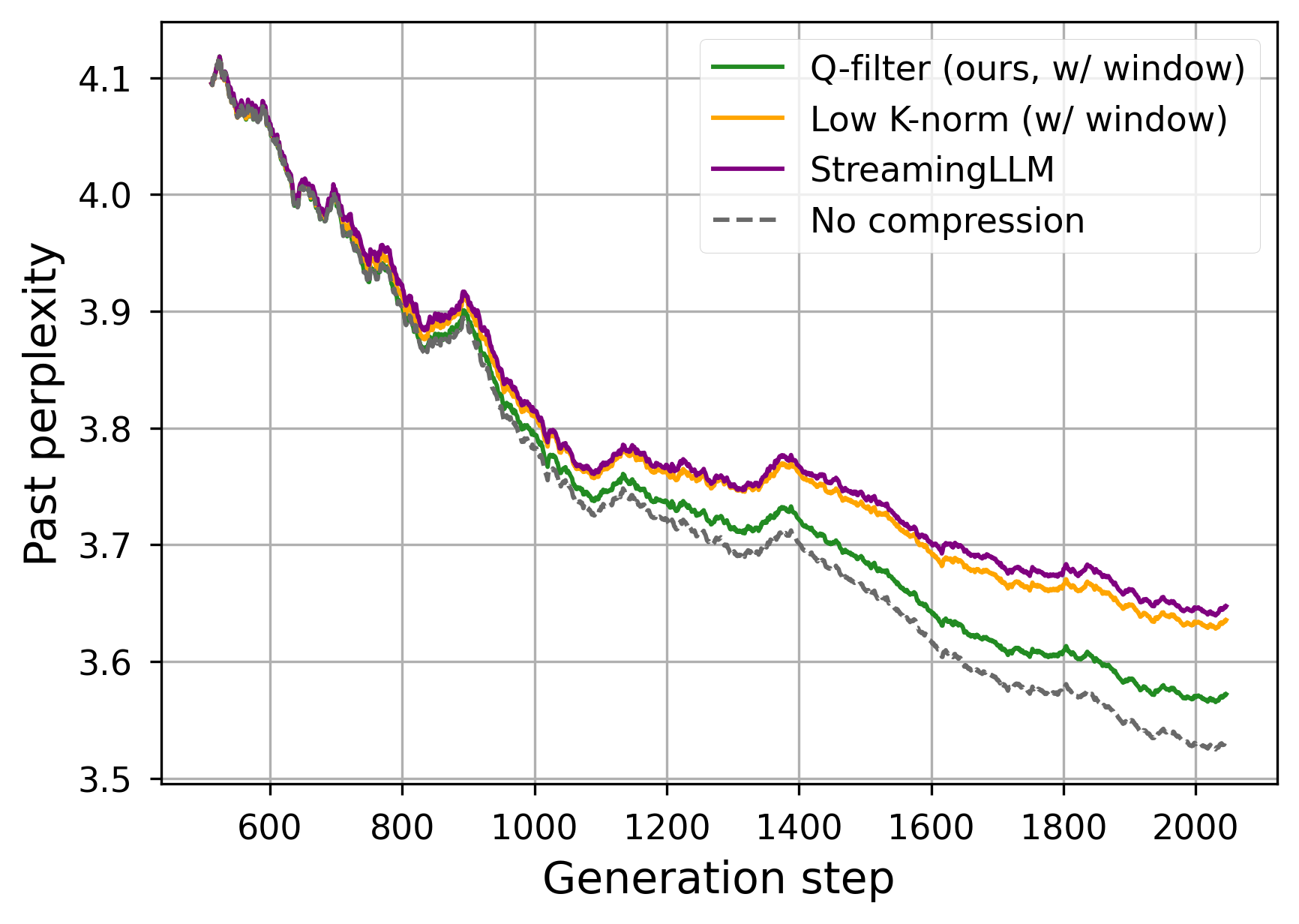}
         \label{fig:stream_70B}
     \end{subfigure}
    \caption{Generation performance for a \kvcache size limited to 512 items for Llama-3.1-8B (top) and Llama-3.1-70B (bottom).}
    \label{fig:stream_perf}
\end{figure}

\paragraph{Language Modelling} To evaluate the performance of \qfilters in the language modelling setup, we perform generation on the Pile dataset \cite{pile}. We let the \kvcache grow up until a certain threshold, after which we start evicting the KV pairs so that the total size never exceeds the maximum threshold. 
We measure performance by tracking the model perplexity computed on past tokens in 20 sequences. We report results for a maximum \kvcache size of 512 pairs in \Cref{fig:stream_perf}.
We observe that \qfilters consistently achieves the lowest perplexity among compression schemes, even for very long contexts. This observation scales to the 70B model, where \qfilters significantly reduces the perplexity gap.
%
This improvement is more pronounced in the latter portions of the sequences, suggesting better retention of relevant contextual information.
\paragraph{Needle in a Haystack}
The Needle-in-a-Haystack task embeds a key piece of information (the “needle”) within a long sequence of distractors (the “haystack”), followed by a question that requires retrieving the needle. This evaluates the model’s ability to handle long-range dependencies and tests how well \kvcache compression retains critical information. If important KV pairs are evicted, the model fails to answer correctly.

\begin{figure}[t]
    \centering
    \begin{subfigure}{0.75\columnwidth}
         \includegraphics[width=\linewidth]{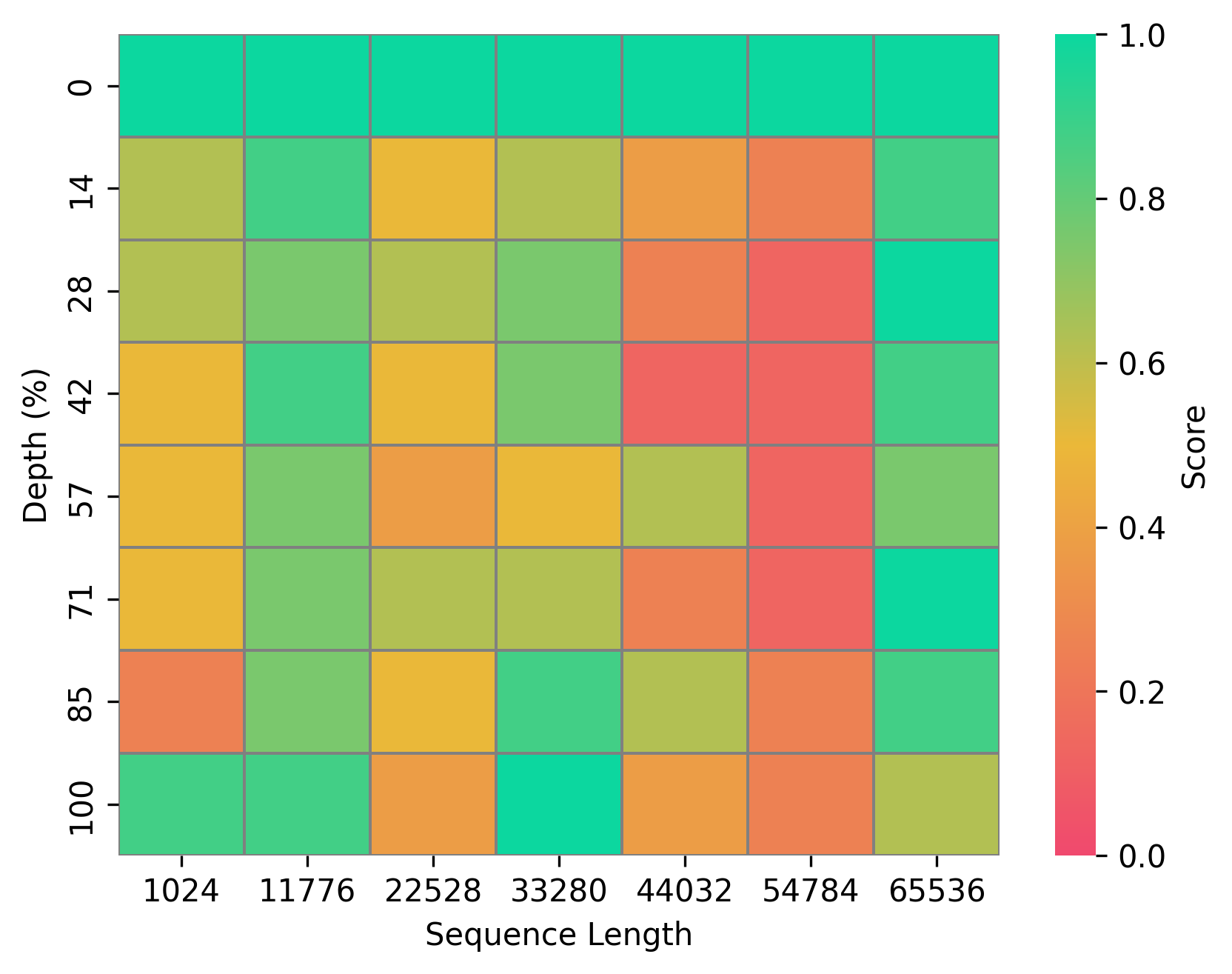}
         \caption{K-norm (average accuracy: 63\%)}
         \label{fig:knorm_needle}
    \end{subfigure}
    \begin{subfigure}{0.75\columnwidth}
         \includegraphics[width=\linewidth]{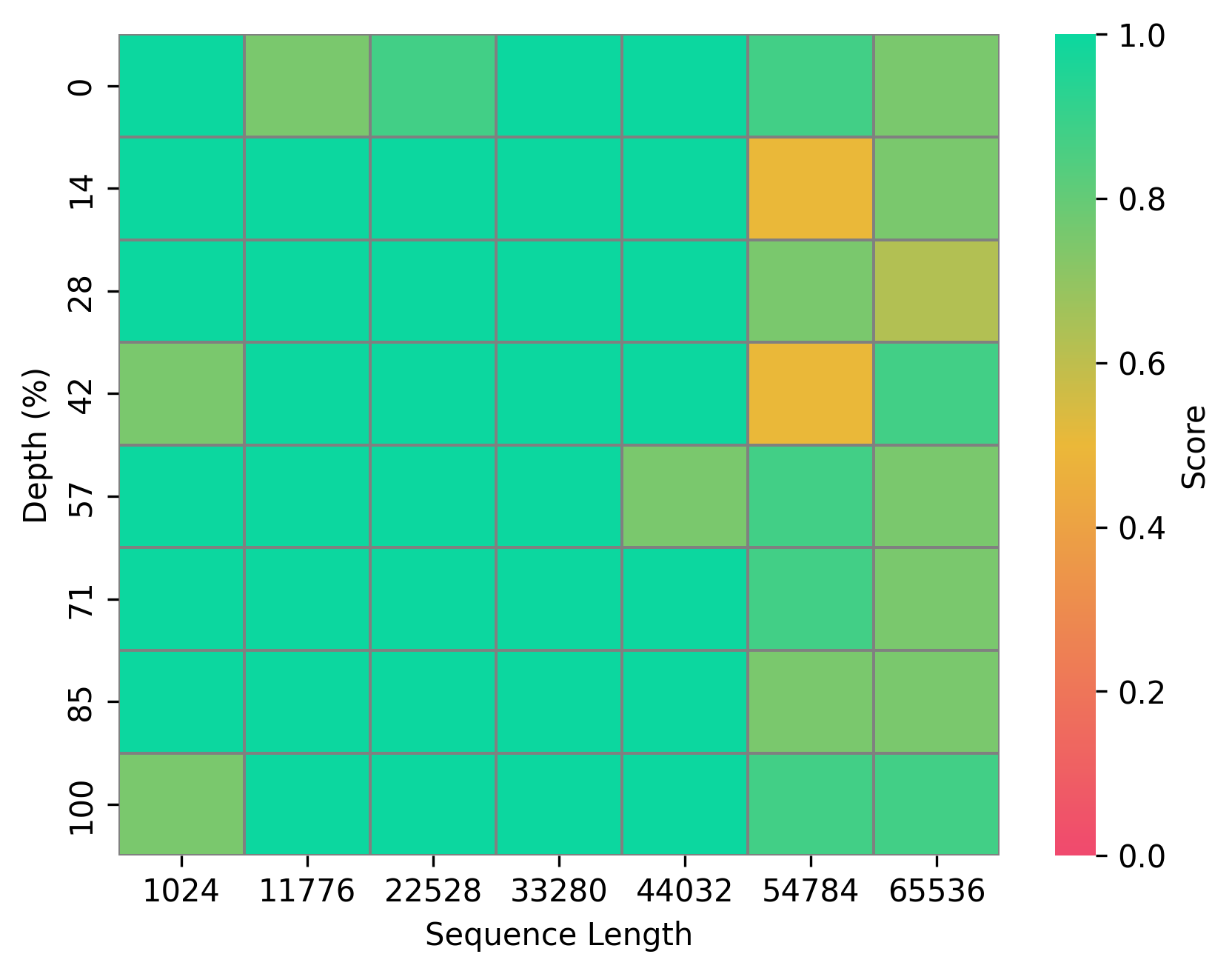}
         \caption{Q-filters (average accuracy: 91\%)}
         \label{fig:qfilter_needle}
    \end{subfigure}
    \caption{Needle-in-a-haystack performance for Llama-3.1-8B using 64x \kvcache compression.}

    \label{fig:niah-general}
\end{figure}

\begin{figure}[t]
\centering
\begin{subfigure}[h]{0.75\columnwidth}
         \includegraphics[width=\textwidth]{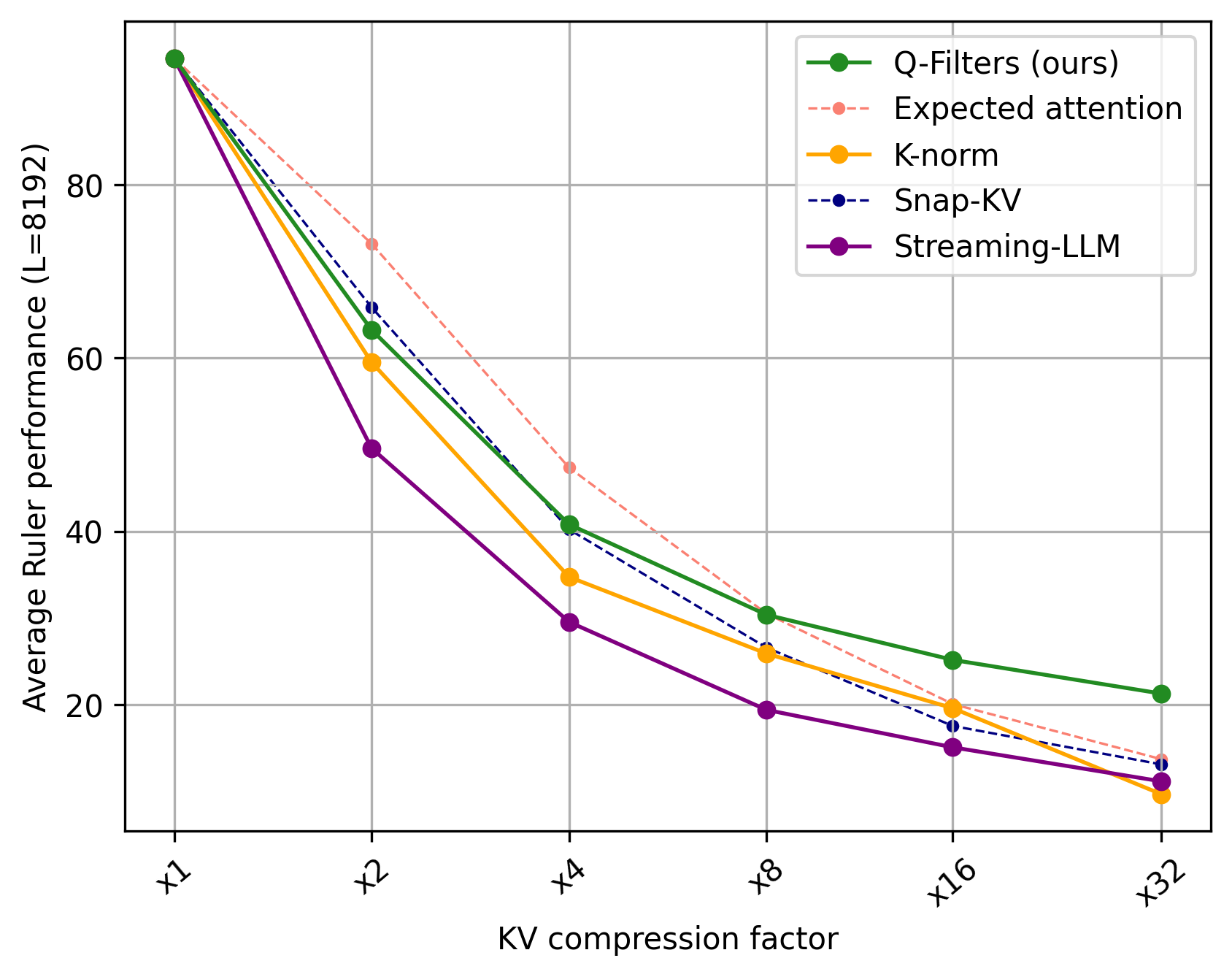}
         \caption{Average performance on Ruler (8192)}
         \label{fig:avg_ruler}
\end{subfigure}
\begin{subfigure}[h]{0.75\columnwidth}
         \includegraphics[width=\textwidth]{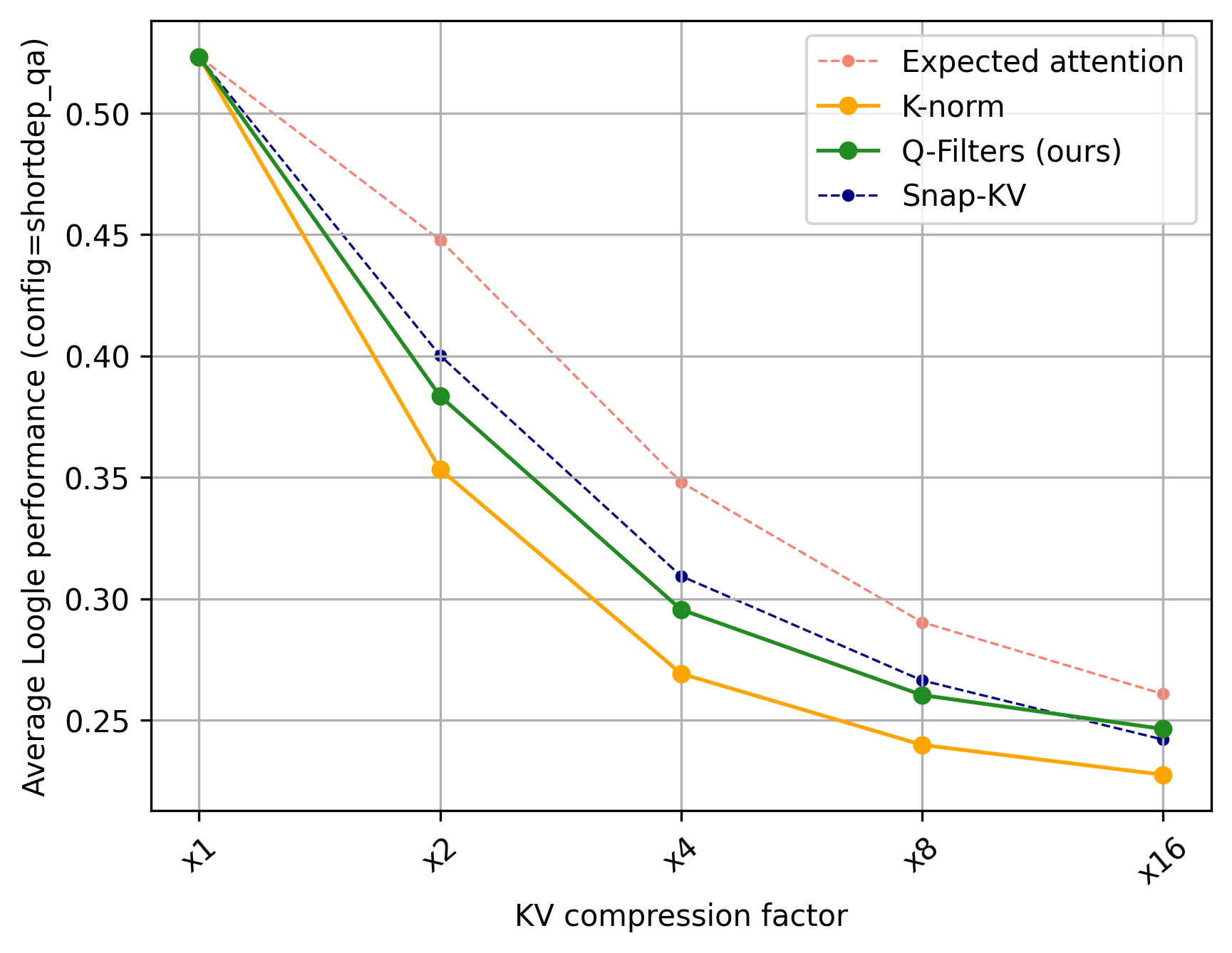}
         \caption{Average performance on Loogle (Short Dependency QA)}
         \label{fig:avg_loogle}
\end{subfigure}
\caption{Average score for different long-context benchmarks using Llama-3.1-8b with different methods and compression ratios}
\label{fig:avg_retrieval}
\end{figure}

We evaluate \qfilters by placing the needle at depths from 1k to 64k tokens and measuring retrieval accuracy. Similarly to \citep{devoto-etal-2024-simple}, we do not compress key-value pairs in the first two layers of the models in this experiment. As shown in \Cref{fig:niah-general}, \qfilters outperforms K-Norm~\citep{devoto-etal-2024-simple}, preserving crucial information even in extremely long contexts.
\paragraph{Ruler Tasks}
We evaluate the proposed method on the Ruler dataset \cite{hsieh2024ruler}, which comprises several sub-tasks that test the model long context modelling abilities, including Multi-hop Tracing, Long Context Aggregation, Long Context Retrieval and Question Answering. The dataset offers 3 variants with different sequence lengths:  4096, 8192, and 16384.
We compare the score on Ruler with several other \kvcache compression methods and show average results in \Cref{fig:avg_ruler}. We report detailed per-task results in \Cref{tab:ruler_detailed} and in \Cref{app:ruler_results}.
We test the model's score for several compression factors ranging from 2$\times$ to 32$\times$. While for some lower compression factors, we find performance on par with other methods, \qfilters achieve the highest score with the strongest compression factor of 32$\times$, demonstrating the method's effectiveness at high compression rates.
%
\begin{table*}[ht]
\centering
\scalebox{0.8}{
\begin{tabular}{@{}rc|rrrrrrrr|c@{}}
\toprule
\textbf{Compression method} &  \textbf{FA-compatible} & \textbf{CWE}  & \textbf{FWE}  & \textbf{Multi-Key} & \textbf{Multi-Query} & \textbf{Multi-Value} & \textbf{Single} & \textbf{QA}   & \textbf{VT}  & \textbf{Average} \\ \midrule
SnapKV                       & \xmark & \textbf{88.7} & {\ul 89.0}      & {\ul 15.1}         & 29.6                 & 28.8                 & {\ul 68.7}      & 42.8          & 83.2         & {\ul 50.5}       \\
Expected Attention           & \xmark                         & 70.0            & 79.3          & 12.0                 & \textbf{59.7}        & {\ul 37.8}           & 31.2            & {\ul 44.2}    & {\ul 96.3}   & 43.2             \\ \midrule
Streaming-LLM                 & \checkmark                       & 53.8          & \textbf{93.4} & 14.1               & 16.8                 & 16.7                 & 15.7            & \textbf{62.3} & 15.8         & 31.6             \\
K-Norm                          & \checkmark                   & 22.9          & 74.8          & 8.7                & 16.6                 & 25.8                 & 55.9            & 20.6          & 32.0           & 31.3             \\
Q-Filters (ours)              & \checkmark                     & {\ul 82.5}    & 80.2          & \textbf{22.9}      & {\ul 49.1}           & \textbf{60.6}        & \textbf{71.1}   & 37.6          & \textbf{100} & \textbf{56.1}    \\ \bottomrule
\end{tabular}}
\caption{Results on the Ruler-4096 dataset for Llama-3.1-70B-Instruct with an 8$\times$ compression ratio. The second column indicates compatibility with FlashAttention. \label{tab:ruler_detailed}}
\end{table*}
\subsection{Robustness of the Calibration Dataset}
\label{sec:calib_dataset}
\begin{figure}[ht]
     \centering
     \includegraphics[width=0.86\columnwidth]{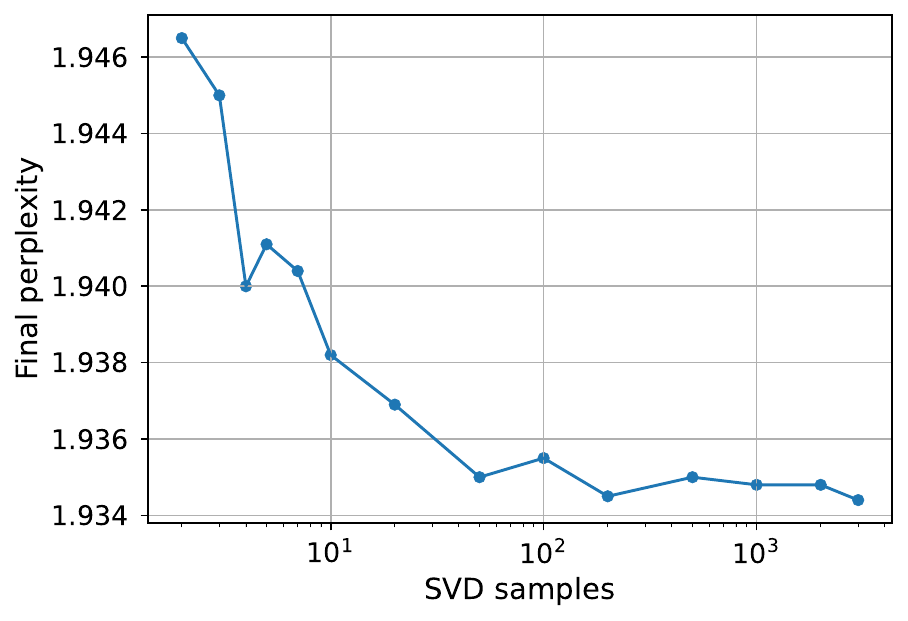}
    \caption{Perplexity after 1024 tokens for Q-Filters obtained using different sizes of $Q^h$ (\cref{eq:qfilters}) to calculate the SVD.}
    \label{fig:calibration_dataset}
\end{figure}

\begin{figure}[ht]
     \centering
     \includegraphics[width=0.9\columnwidth]{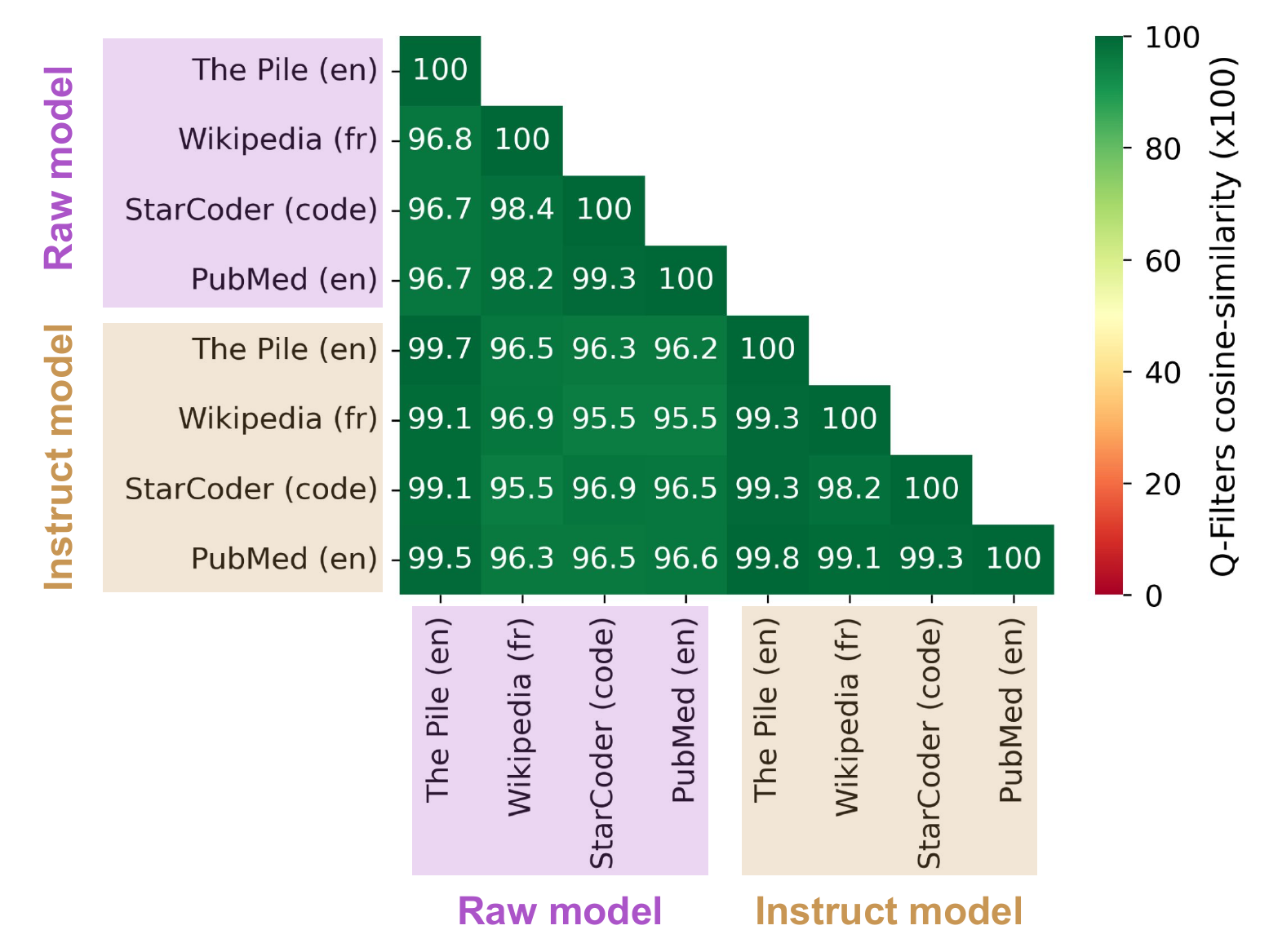}
     \caption{Cosine-similarity between Q-Filters computed on datasets coming from different domains and languages and on pre-trained and post-trained models. The scores are averaged over all layers and heads.}
     \label{fig:score_mat}
\end{figure}

In \Cref{fig:calibration_dataset}, we analyse how the calibration dataset size impacts the performance of our \qfilters computation. Our experimental results demonstrate that increasing the number of samples in the calibration dataset leads to an improvement in average perplexity, although the marginal benefits diminish beyond a certain point, namely around 1k samples.
This suggests that while larger calibration datasets generally produce more robust Q-Filters, there exists a practical trade-off balancing computational cost and performance benefits.
Based on these empirical findings and computational efficiency considerations, we standardized our experimental protocol to utilize 3,000 samples for computing the Q-Filters across all subsequent experiments. %
Another important consideration in the development of robust \qfilters is the choice of calibration dataset. To investigate this aspect, we conducted a systematic analysis using multiple diverse datasets and model versions in \Cref{fig:score_mat}. Our experiments revealed that the Q-Filter vectors exhibit remarkable stability across different calibration datasets, with a high average cosine similarity between vectors computed from distinct sources.
This finding suggests that our method is relatively insensitive to the specific choice of calibration data, provided it maintains sufficient diversity and quality. Based on these results, we opted to use a carefully curated subset of the Pile dataset \citep{pile} for all Q-Filter computations.
\subsection{\qfilters Estimation Overhead}
It could be argued that our method introduces a memory overhead as we need to store the Q-Filters on-device. Nevertheless, for a model using $l$ layers and $n_H$ heads, storing the Q-Filters requires $l \times n_H \times d_H$ parameters. For Llama-3.2-1B, this is 36k$\times$ smaller than the total parameter count and 196k$\times$ smaller in the case of Llama-3.2-405B.
Another source of overhead could be attributed to the initial computation of the filters that are required for every new model.
We find that passing 20 samples of length 2048 through the model and performing the SVD on 3k randomly sampled representations for each head is sufficient to obtain strong performance. In our experiments with Llama-3.2-70B, computing the filters took less than 3 minutes on two A100-80GB GPUs. 
This cost is thus negligible when compared with the cost of inference.
\subsection{Throughput and Scalability}
In this section, we analyze the time and memory overhead induced by the \qfilters method.
Our approach is more efficient than many \kvcache compression methods, as it estimates the relevance of a $K^h$ representation \textit{without materializing the attention maps}. This property makes it compatible with memory-efficient self-attention implementations such as FlashAttention \citep{dao2023flashattention2}.  
During inference, \qfilters maintains the same theoretical time complexity as the $K$-norm method \citep{devoto-etal-2024-simple}, since computing a norm and a scalar product require a comparable number of floating-point operations.  

\begin{figure}[t]
      \centering
      \includegraphics[width=0.86\columnwidth]{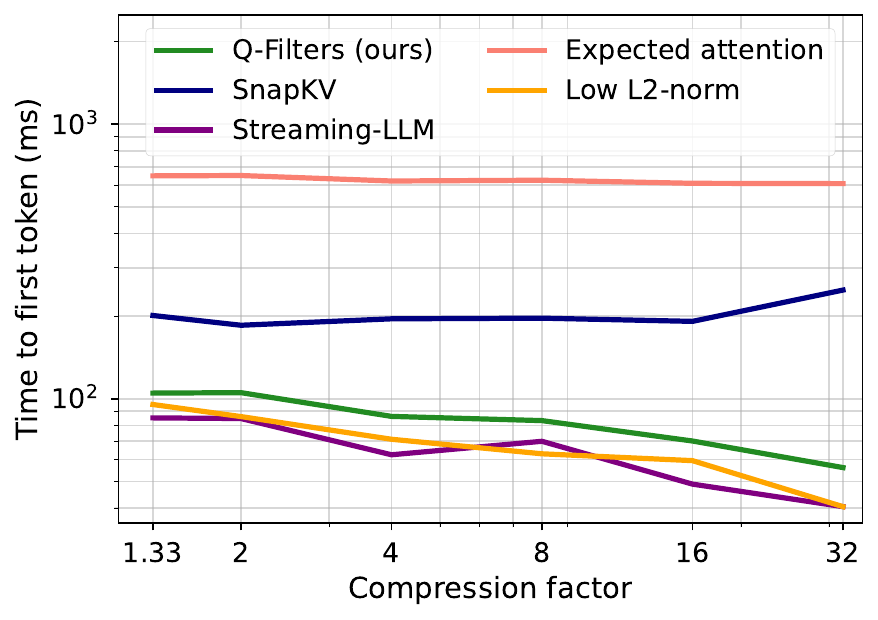}
     \caption{First token latency across KV Cache compression methods of Llama-3.2-8B with a length of 64k prompt. 
     }
     \label{fig:first_token_across_seq}
\end{figure}

By avoiding the explicit computation of attention scores, our method achieves lower inference latency compared to existing approaches. To quantify this efficiency, we measure the \textit{Time to First Token} across different methods in \Cref{fig:first_token_across_seq}.  Time to First Token (TTFT) refers to the latency between submitting a prompt and receiving the first generated token. This metric is particularly relevant in scenarios where fast response times are critical, such as interactive AI applications. Compressing the \kvcache directly impacts TTFT: by reducing the memory footprint of the \kvcache, it allows a larger portion of the prompt context to fit within fast-access memory, minimizing memory swapping overhead. As a result, compression techniques that efficiently manage the \kvcache should significantly reduce initial response latency.  Notably, our experiments show that Q-Filters maintain this performance advantage even as the sequence length increases, suggesting better scalability compared to methods that require explicit attention computation.
\section{Limitations}
In \Cref{app:gen_res}, we run generation experiments on Qwen-2.5-7B-Instruct \citep{qwen2025qwen25technicalreport}, and we observe that, although the results still favour the \qfilters method, the gap is less clear compared to the Llama models.
Our main hypothesis for this discrepancy lies in the slightly different attention mechanism used in Qwen-2.5 suite, which adds a bias to the QKV projection. Hence, it is likely that the geometrical observations made in \cref{sec:method} are not accurate in that case. Similarly, initial experiments with Olmo-2 models \citep{olmo20252olmo2furious} were unsuccessful, which can be explained by their use of the QK-normalization technique \citep{pmlr-v202-dehghani23a}. These different tricks would most likely require an adaptation of our analysis to yield a better approximation of the attention distributions.
\section{Related Works}
After the success of long-context models~\citep{reid2024gemini, anthropic2024claude, achiam2023gpt}, compressing the \kvcache has become a key research focus to enable processing of long-context inputs. 
Some methods reduce the \kvcache size by modifying the model architecture. For example, \citet{ainslie2023gqa} and \citet{shazeer2019fasttrans} reuse the same Keys for multiple queries, thereby reducing redundancy in storage. \citet{nawrot2024dynamic} propose a dynamic token-merging strategy, learning which KV pairs to merge.
While these approaches achieve significant compression, they require training or fine-tuning, making them less practical in real-world scenarios where retraining the model from scratch is not feasible. In contrast, our method requires only a short, computationally inexpensive calibration step, avoiding parameter updates entirely.
Recently \citet{deepseekaiv2} introduced a Multi-Head Latent Attention, a modification to the standard attention mechanism that performs a low-rank reduction of the \kvcache during pre-training.
Training-free approaches aim to compress the \kvcache without modifying the model, typically by approximating the attention score over long sequences and prioritizing tokens with higher importance. Among these, \citet{xiao2024efficient} focus on language modelling tasks and propose always retaining the first token(s) (as an attention sink) and the last $n$ tokens in a sliding window. Also, \citet{h2o} focuses on generation tasks and introduces a policy that evicts tokens during generation based on a scoring function derived from cumulative attention. 
In contrast, other works focus on the task of compressing a large prompt provided by the user. \citet{li2024snapkv} uses attention from the last part of the prompt to estimate KV pairs importance. With the same goal, \citet{cai2024pyramidkvdynamickvcache} assigns more cache budget to lower layers and less to higher layers. Finally, \citet{vatp} proposes to rescale the KV score of other methods by the $L_1$ norm of the Values.
In contrast, our approach is not tailored to a specific use case but provides competitive performance across both synthetic tasks and real-world scenarios, including in-context learning and chat-based interactions. 
Additionally, many of these approaches are incompatible with FlashAttention \citet{dao2023flashattention2} due to their reliance on accessing the full attention weights, which FlashAttention does not expose.
\section{Conclusion}
We introduced \qfilters, a novel training-free method for \kvcache compression. We show that projecting the Key representations on the main SVD component of the Query vectors results in an accurate approximation of the attention scores.
\qfilters is extremely efficient 
and is compatible with FlashAttention as it does not require accessing the attention scores. We validated our method on several tasks (Language modelling, NIAH, Ruler) and models up to 70B parameters, and showed competitive performance with respect to more costly state-of-the-art \kvcache compression methods.
%
%
%
\section{Impact Statement}
%
%
This paper introduces Q-Filters, a training-free technique for compressing the Key-Value cache in large language models by exploiting the geometry of Query and Key vectors.
By discarding less important representations through a single projection direction, Q-filters substantially reduce memory usage while preserving performance across long contexts.
Crucially, it remains compatible with memory-efficient attention mechanisms, facilitating practical adoption in real-world scenarios. This advancement addresses pressing scalability and latency challenges and offers a fresh perspective on harnessing geometrical insights to develop more efficient language modelling strategies.

\section{Acknowledgements}

This collaboration was made possible by my academic visit to Prof. Edoardo Ponti's lab at the University of Edinburgh. I express my sincere gratitude to Prof. Ponti for this opportunity and for our exciting discussions.

This work was funded by the last author's chair in the PRAIRIE institute funded by the French national agency ANR as part of the ``Investissements d'avenir'' programme under the reference ANR-19-P3IA-0001. 

This work was granted access to the HPC resources of IDRIS under the allocation 2024-AD011013680R2 made by GENCI.
%
%


\bibliographystyle{icml2025}

\bibliography{references}

\clearpage
\appendix

\section{Proof of \Cref{thm:main}}
\label{app:proof}
We begin the proof by writing $\langle Q^h_i, K^h_j \rangle$ in the basis $B$:
\begin{equation*}
\begin{split}
    \mathbb{E}_{Q^h_i}(\langle Q^h_i, k\rangle) & = \mathbb{E}_{Q^h_i}(\langle Q^h_i, u^h \rangle)\langle k, u^h \rangle \\ & + \sum_{m=2}^{d_h} \mathbb{E}_{Q^h_i}(\langle Q^h_i, u_m \rangle)\langle k, u_m \rangle
\end{split}
\end{equation*}

\Cref{assum:2} states that $\mathbb{E}_{i, X}(\langle Q^h_i, u_m \rangle) \approx 0$, which lets us do the following approximation:
$$
\sum_{m=2}^{d_h} \mathbb{E}_{Q^h_i}(\langle Q^h_i, u_m \rangle)\langle k, u_m \rangle \approx 0
$$

By combining \Cref{assum:1} and \Cref{assum:2}, we also have that: 
$$
\mathbb{E}_{Q^h_i}(\langle Q^h_i, u^h \rangle) > 0
$$

We conclude the proof by setting $\kappa^h = \mathbb{E}_{Q^h_i}(\langle Q^h_i, u^h \rangle)$.
\section{Generation Results}
\label{app:gen_res}
We compute the final perplexity of Llama-3.1-70B in the memory-constrained setup for various compression factors and methods.
\begin{figure}[h!]
\centering
    \includegraphics[width=0.8\columnwidth]{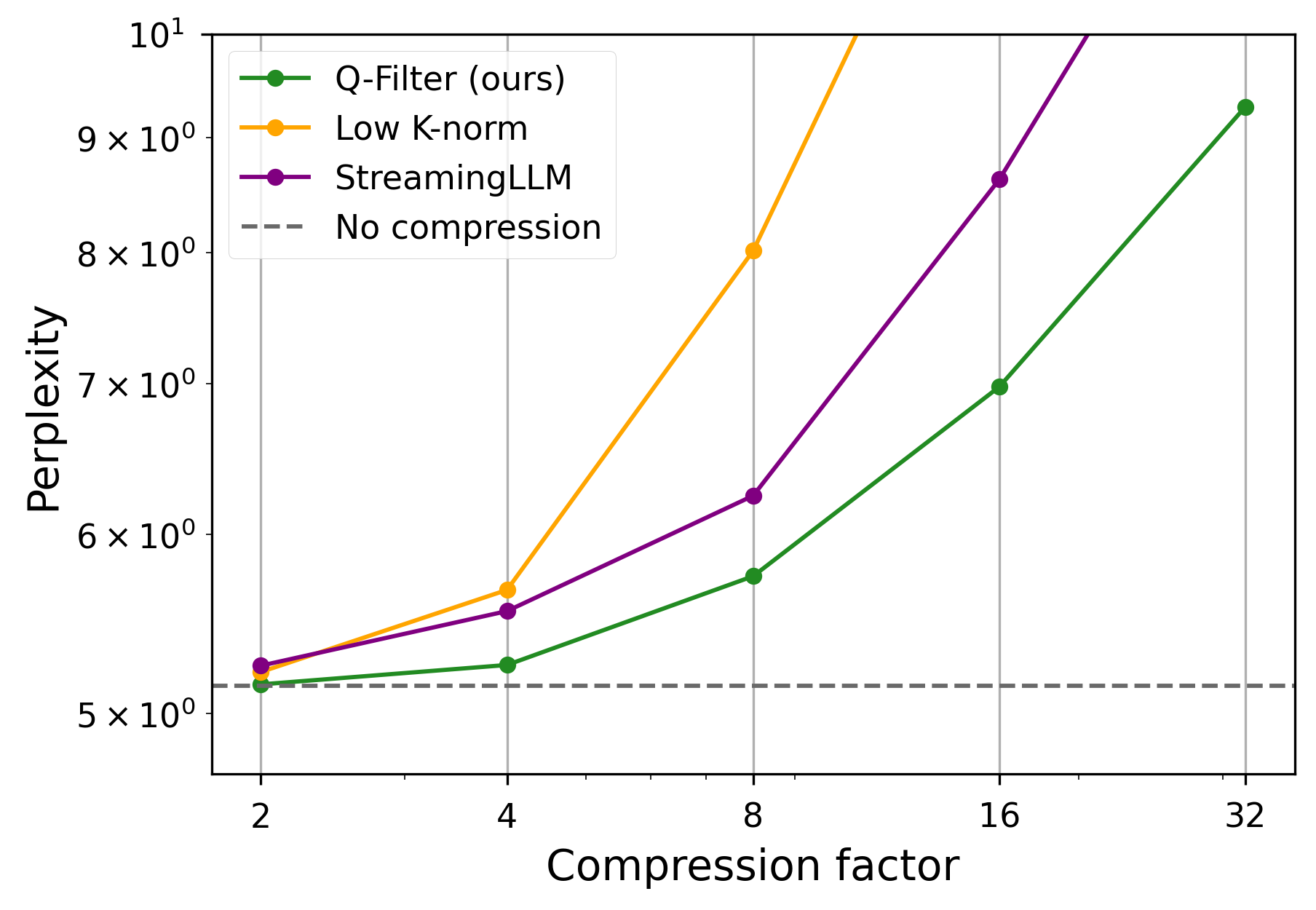}
    \caption{Final perplexity after 512 tokens for Llama-3.1-70B in the memory-constrained generation scenario.}
    \label{fig:final_ppl}
\end{figure}

\begin{figure}[ht]
      \centering
      \includegraphics[width=0.8\columnwidth]{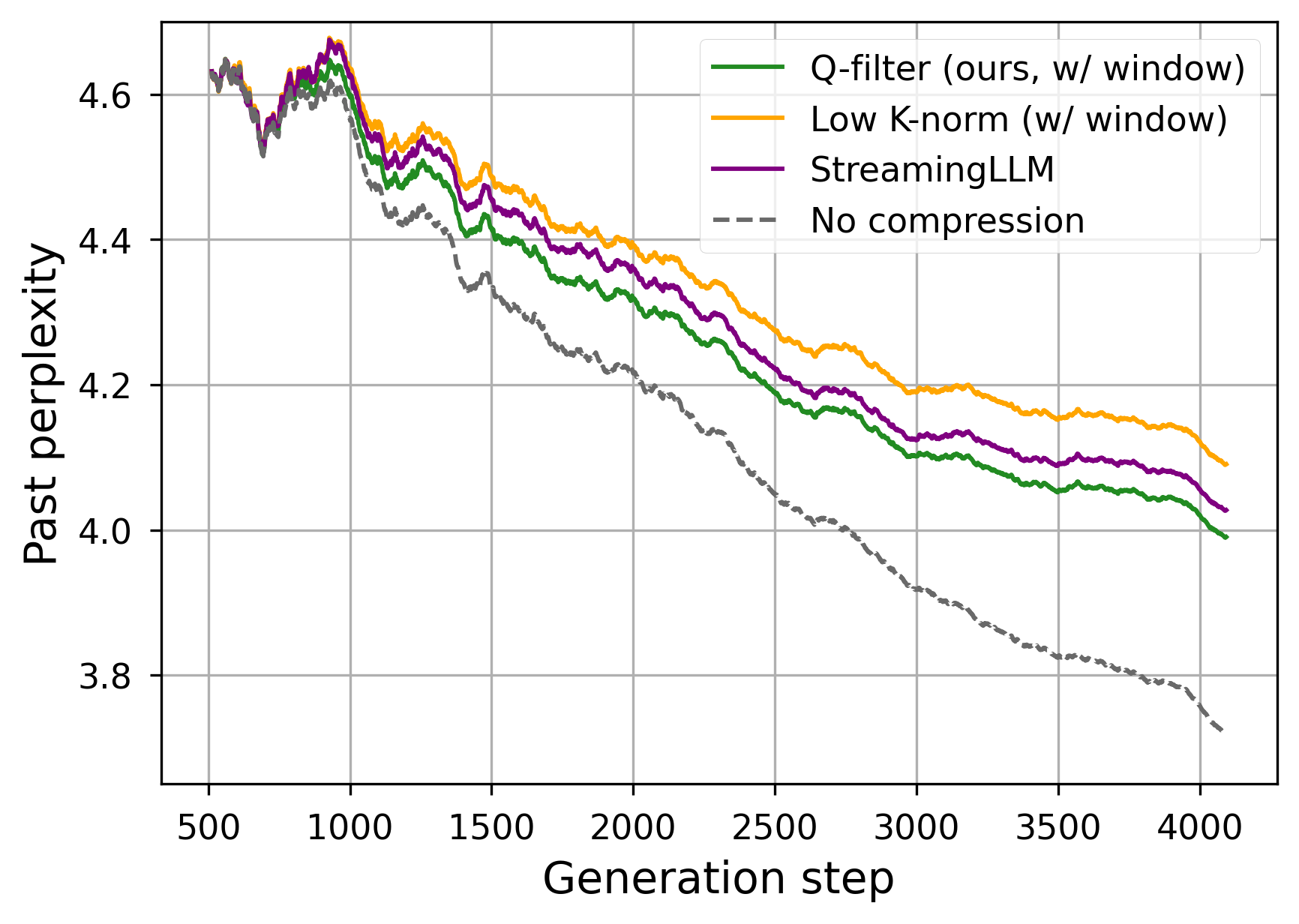}
     \caption{Perplexity of the Qwen-2.5-7B-Instruct model along generation.}
     \label{fig:qwen_gen}
\end{figure}

\begin{figure}[ht]
      \centering
      \includegraphics[width=0.8\columnwidth]{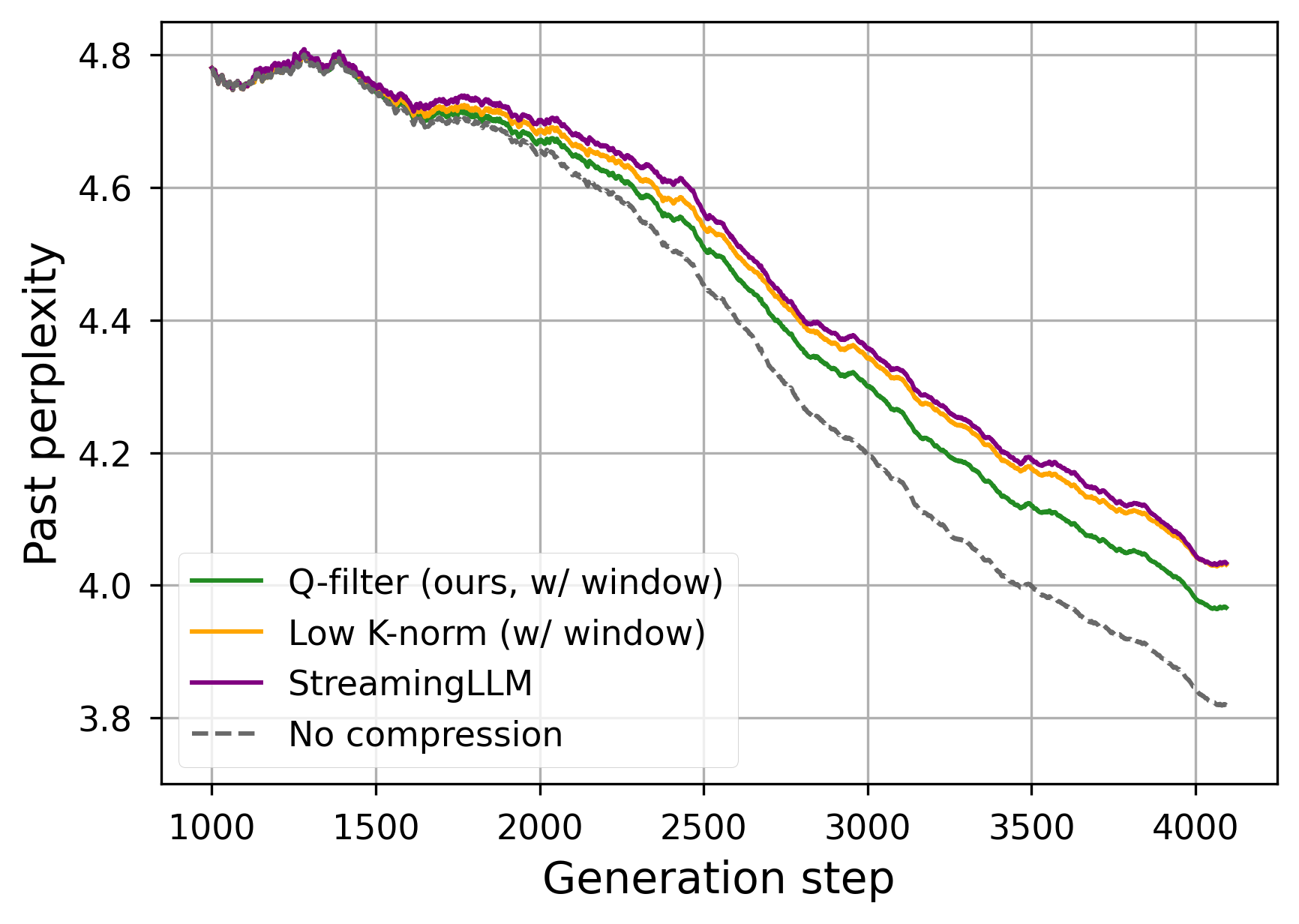}
     \caption{Perplexity of the Llama-3.2-1B model along generation.}
     \label{fig:ll1b_gen}
\end{figure}

We also run a study similar to the one conducted in \Cref{fig:stream_perf} with Qwen-2.5-7B-Instruct, which we display in \Cref{fig:qwen_gen}, and with Llama-3.2-1B, which we display in \Cref{fig:ll1b_gen}.

\section{Ruler Results}
\label{app:ruler_results}

\begin{figure*}[h!]
     \centering
     \begin{subfigure}[h]{0.32\textwidth}
         \centering
         \includegraphics[width=\textwidth]{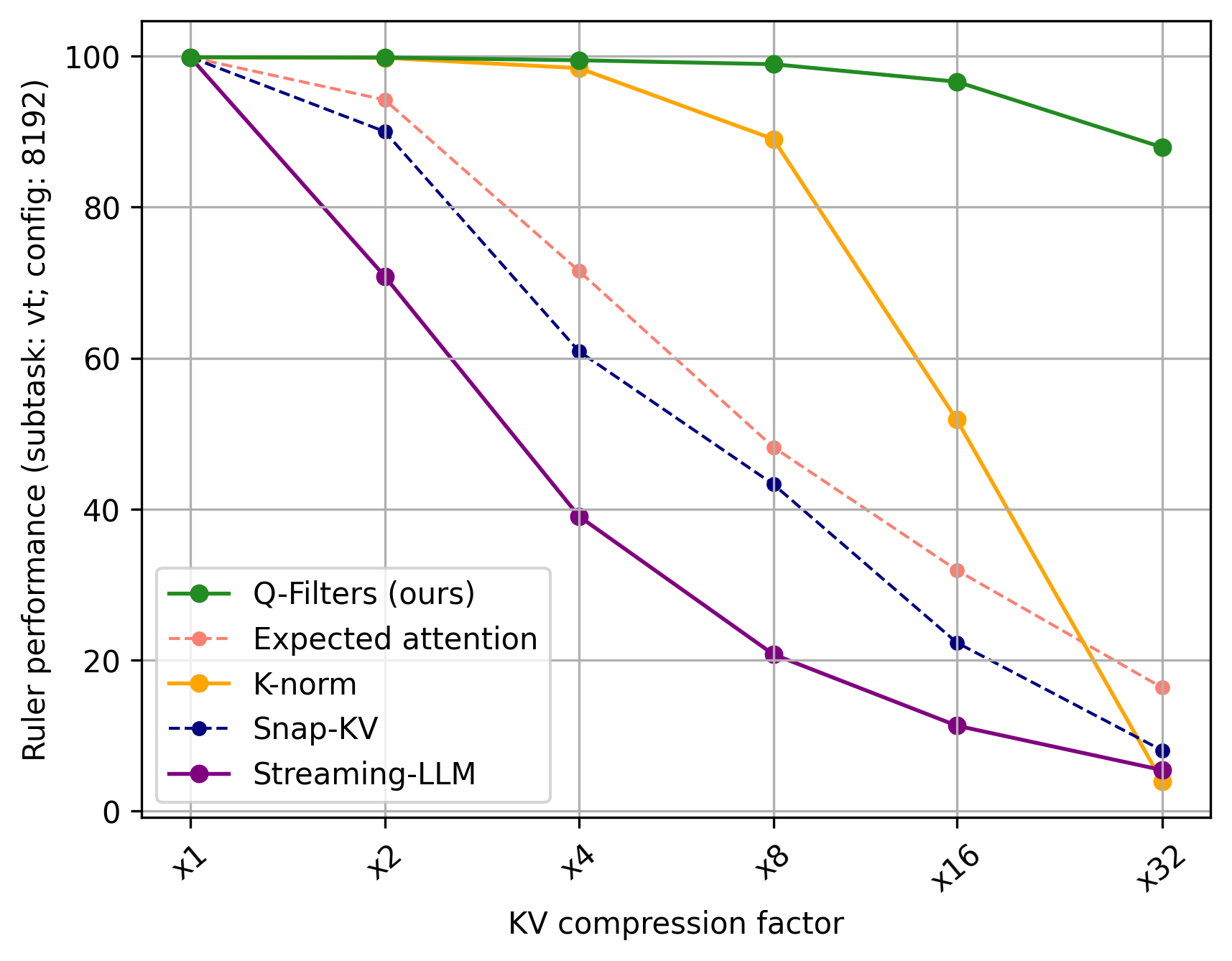}
         \caption{Variable tracking}
         \label{fig:ruler_vt}
     \end{subfigure}
     \begin{subfigure}[h]{0.32\textwidth}
         \centering
         \includegraphics[width=\textwidth]{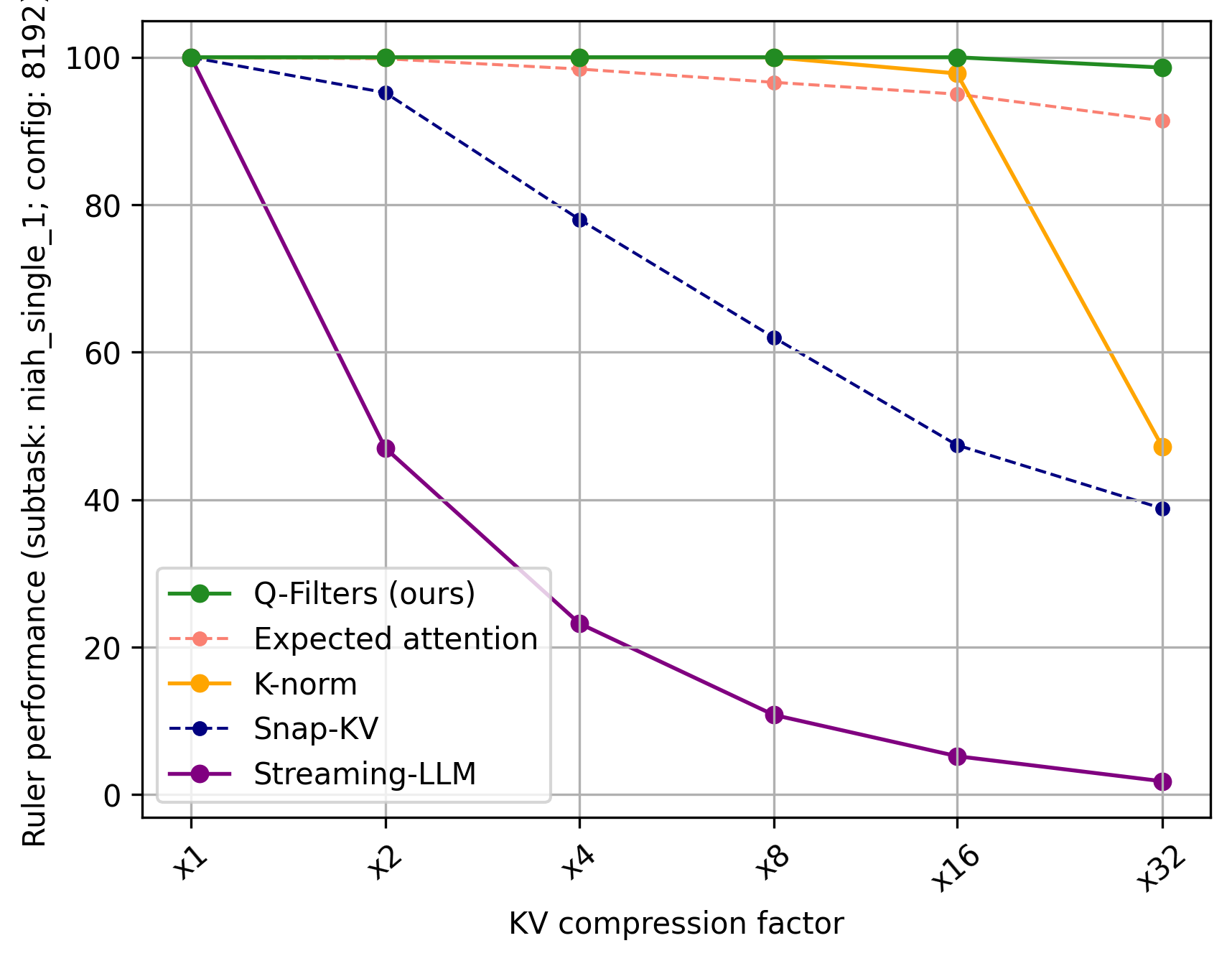}
         \caption{NIAH - single (1)}
         \label{fig:ruler_niah_single_1}
     \end{subfigure}
     \begin{subfigure}[h]{0.32\textwidth}
         \centering
         \includegraphics[width=\textwidth]{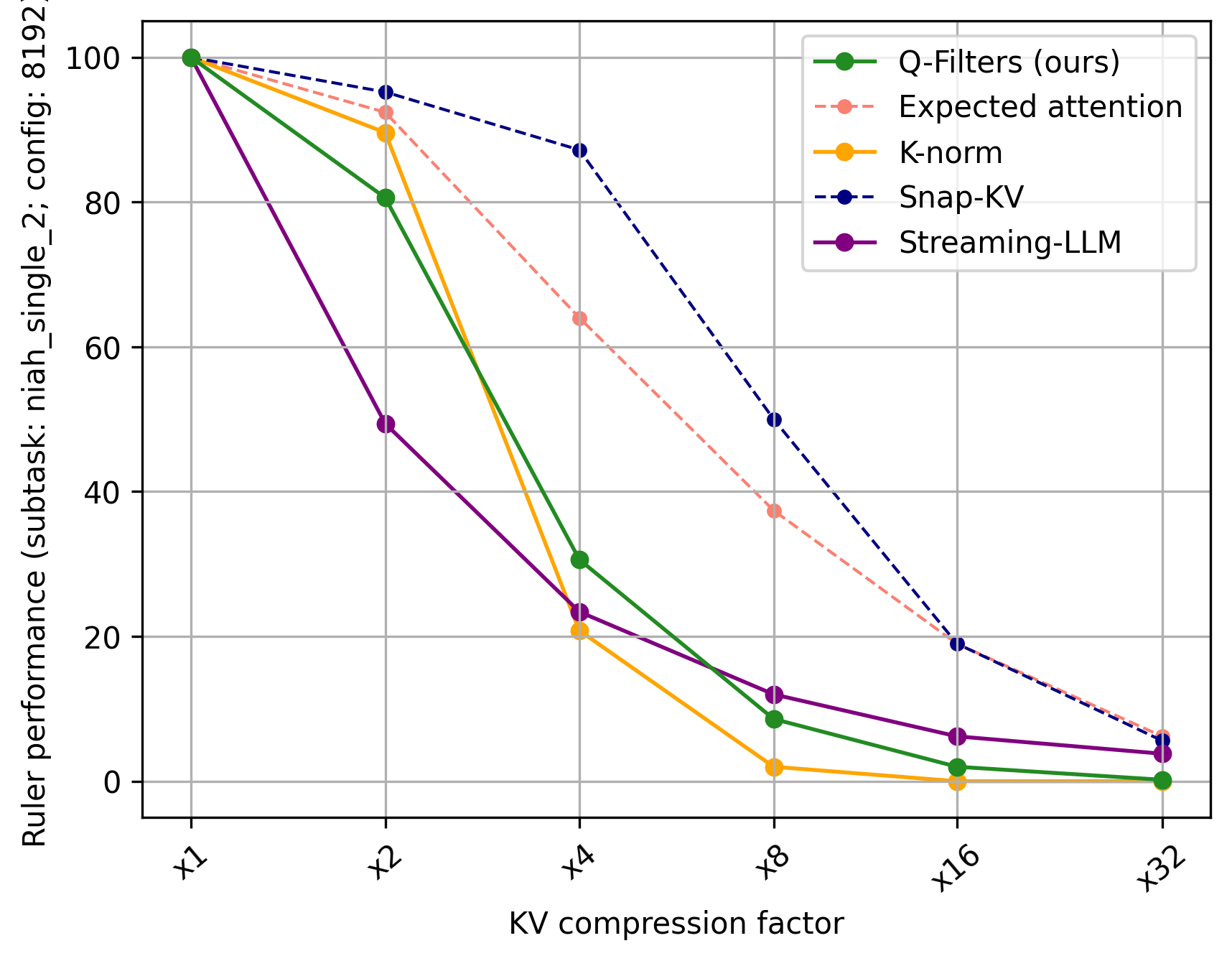}
         \caption{NIAH - single (2)}
         \label{fig:ruler_niah_single_2}
     \end{subfigure}
     \begin{subfigure}[h]{0.32\textwidth}
         \centering
         \includegraphics[width=\textwidth]{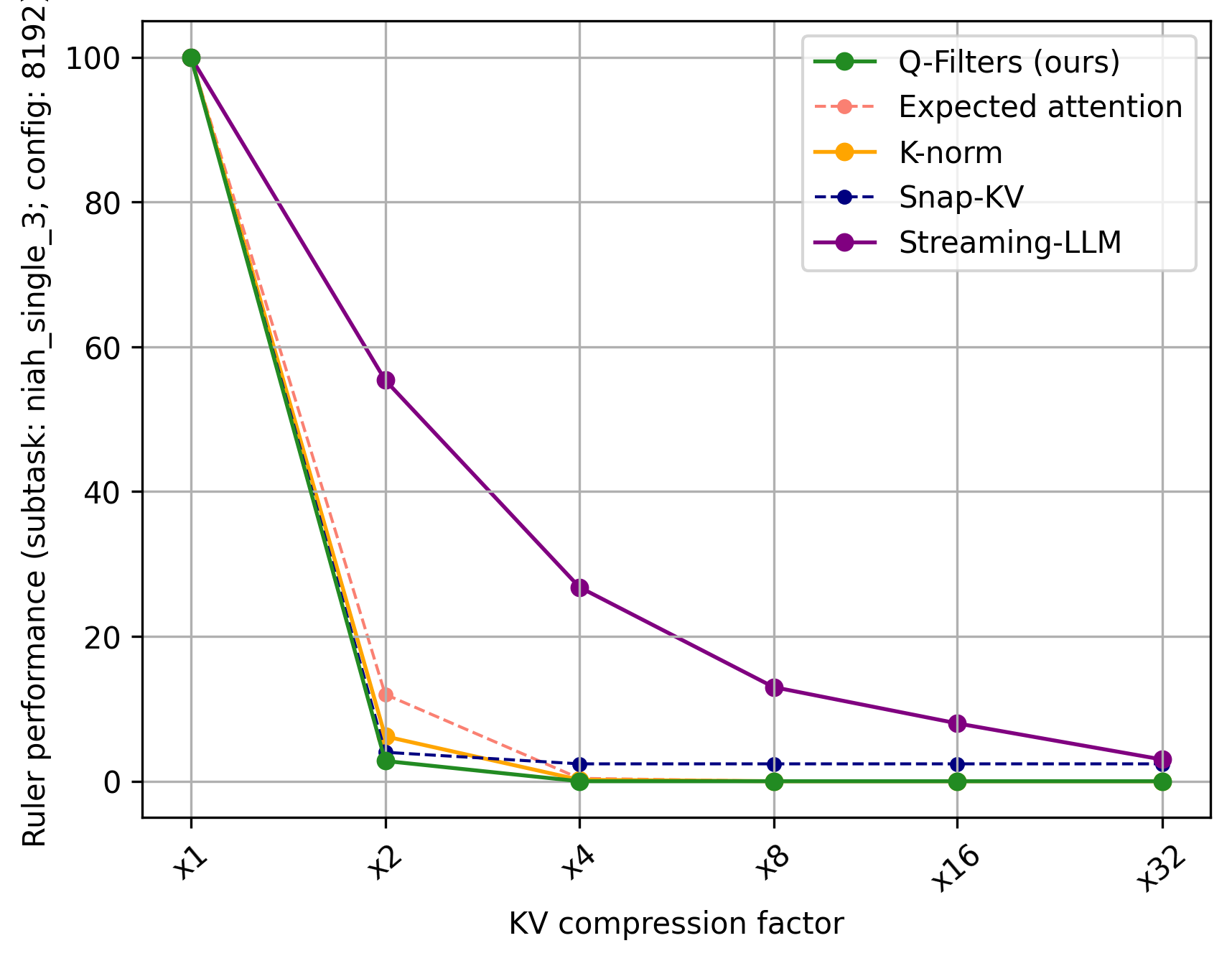}
         \caption{NIAH - single (3)}
         \label{fig:ruler_niah_single_3}
     \end{subfigure}
     \begin{subfigure}[h]{0.32\textwidth}
         \centering
         \includegraphics[width=\textwidth]{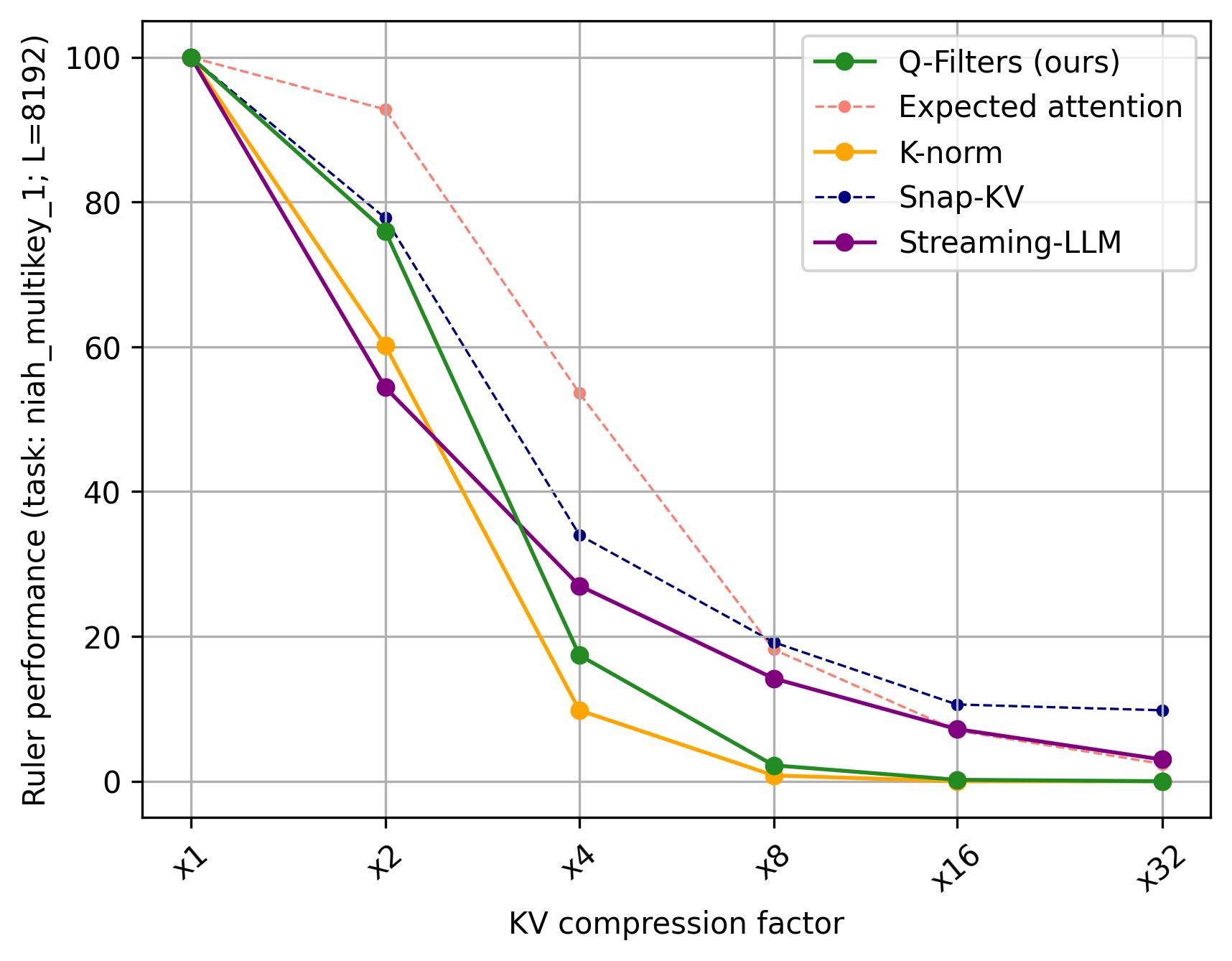}
         \caption{NIAH - Multi-key (1)}
         \label{fig:ruler_niah_single_3}
     \end{subfigure}
     \begin{subfigure}[h]{0.32\textwidth}
         \centering
         \includegraphics[width=\textwidth]{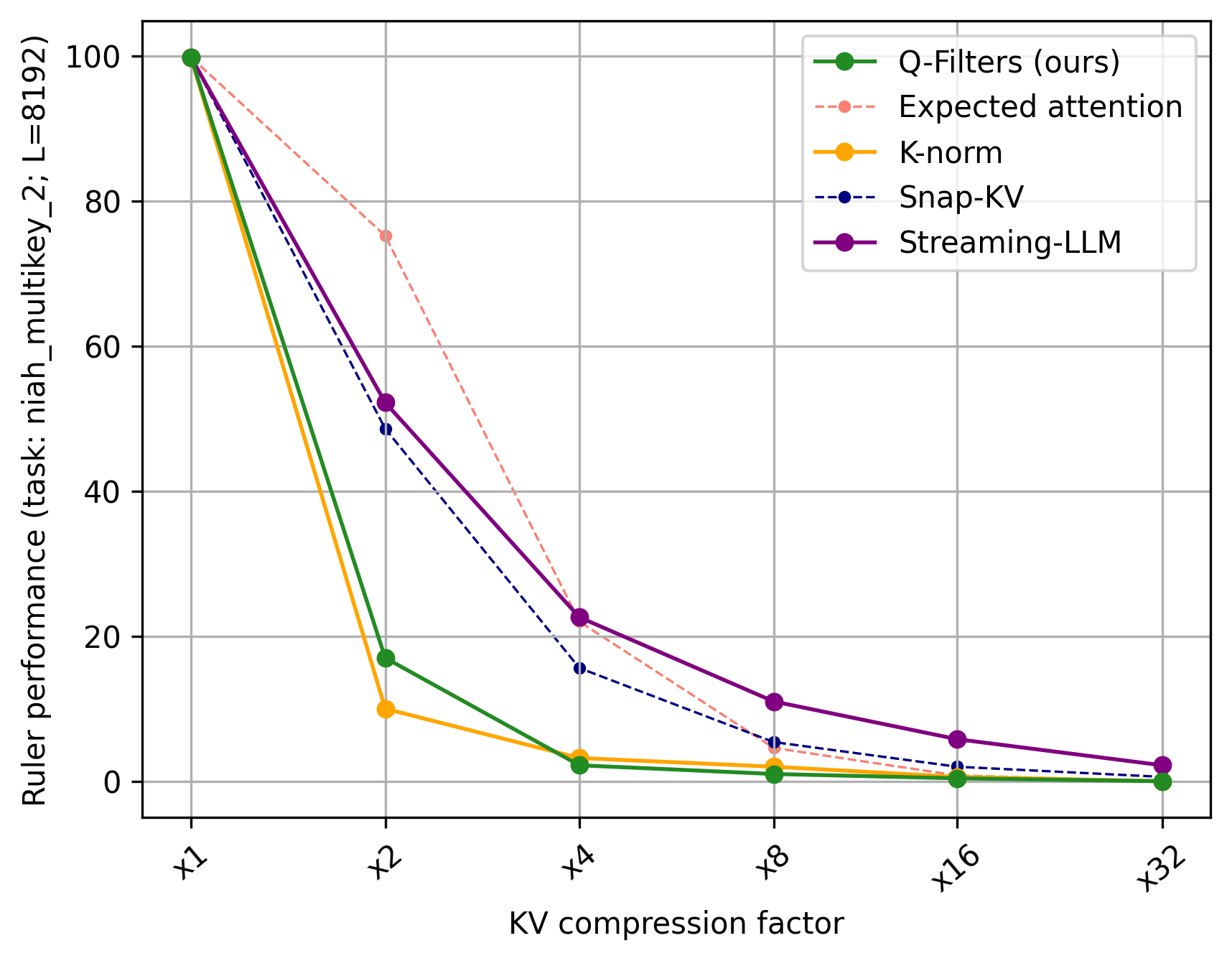}
         \caption{NIAH - Multi-key (2)}
         \label{fig:ruler_niah_single_3}
     \end{subfigure}
     \begin{subfigure}[h]{0.32\textwidth}
         \centering
         \includegraphics[width=\textwidth]{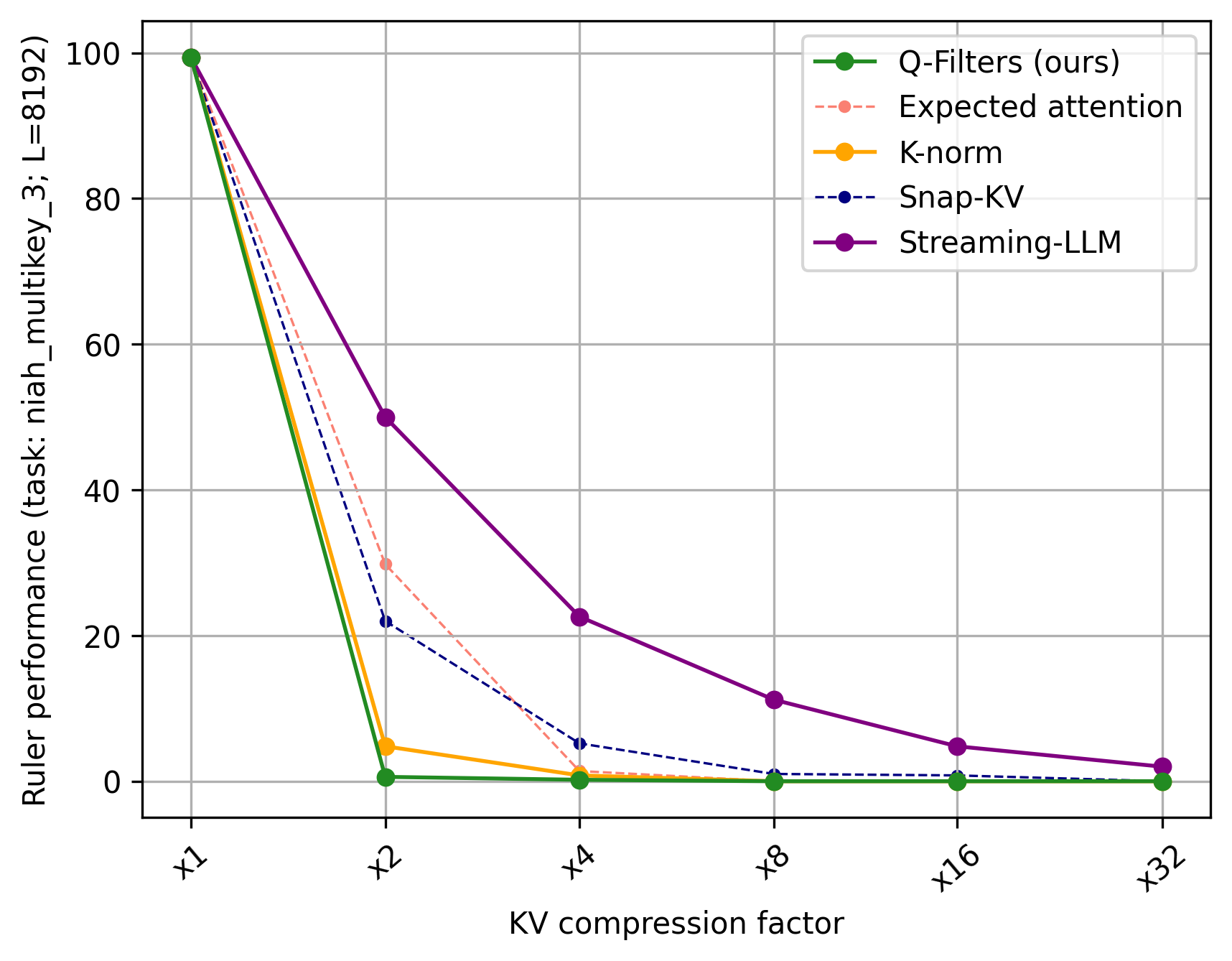}
         \caption{NIAH - Multi-key (3)}
         \label{fig:ruler_niah_single_3}
     \end{subfigure}
     \begin{subfigure}[h]{0.32\textwidth}
         \centering
         \includegraphics[width=\textwidth]{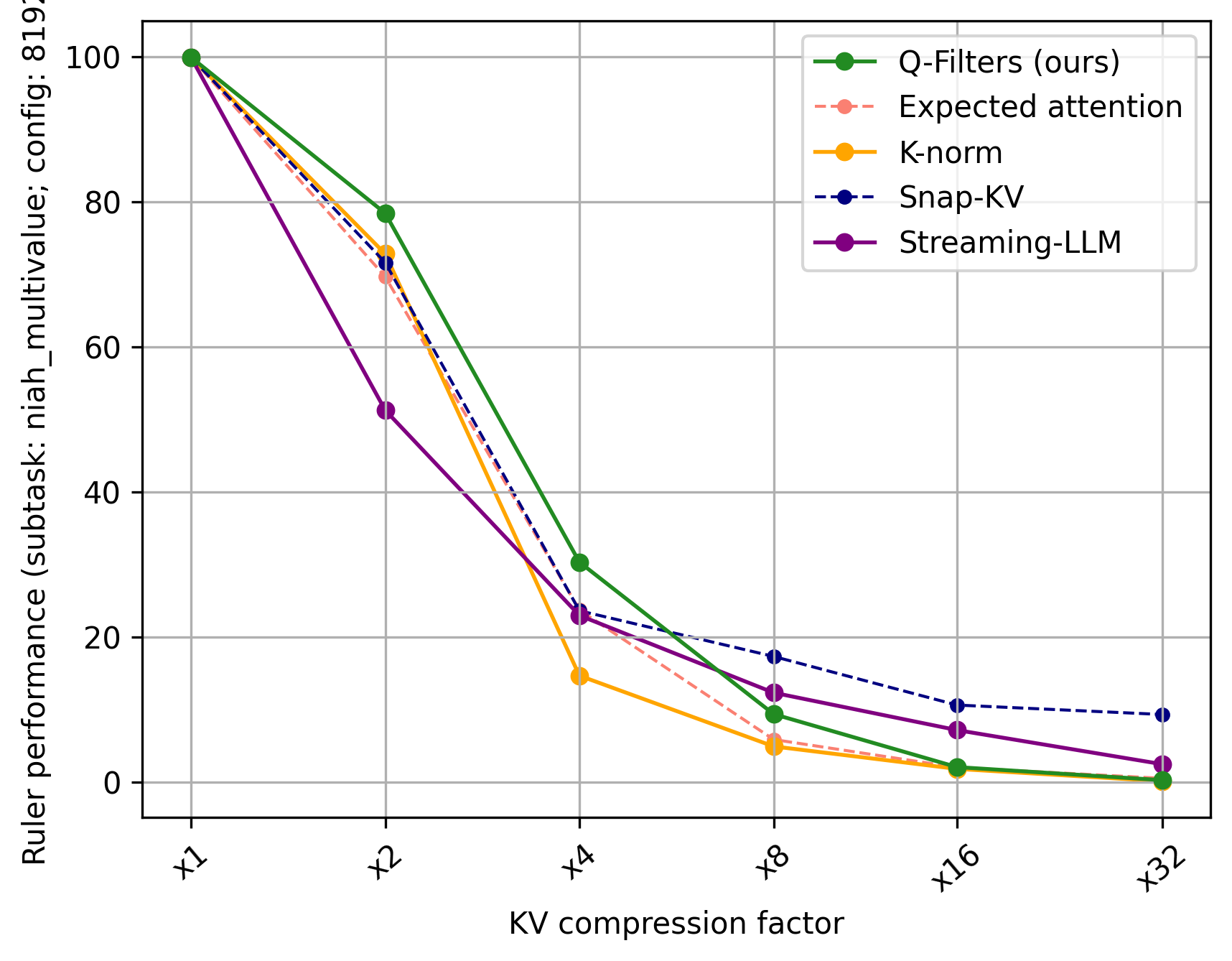}
         \caption{NIAH - Multi-Value}
         \label{fig:ruler_niah_multival}
     \end{subfigure}
     \begin{subfigure}[h]{0.32\textwidth}
         \centering
         \includegraphics[width=\textwidth]{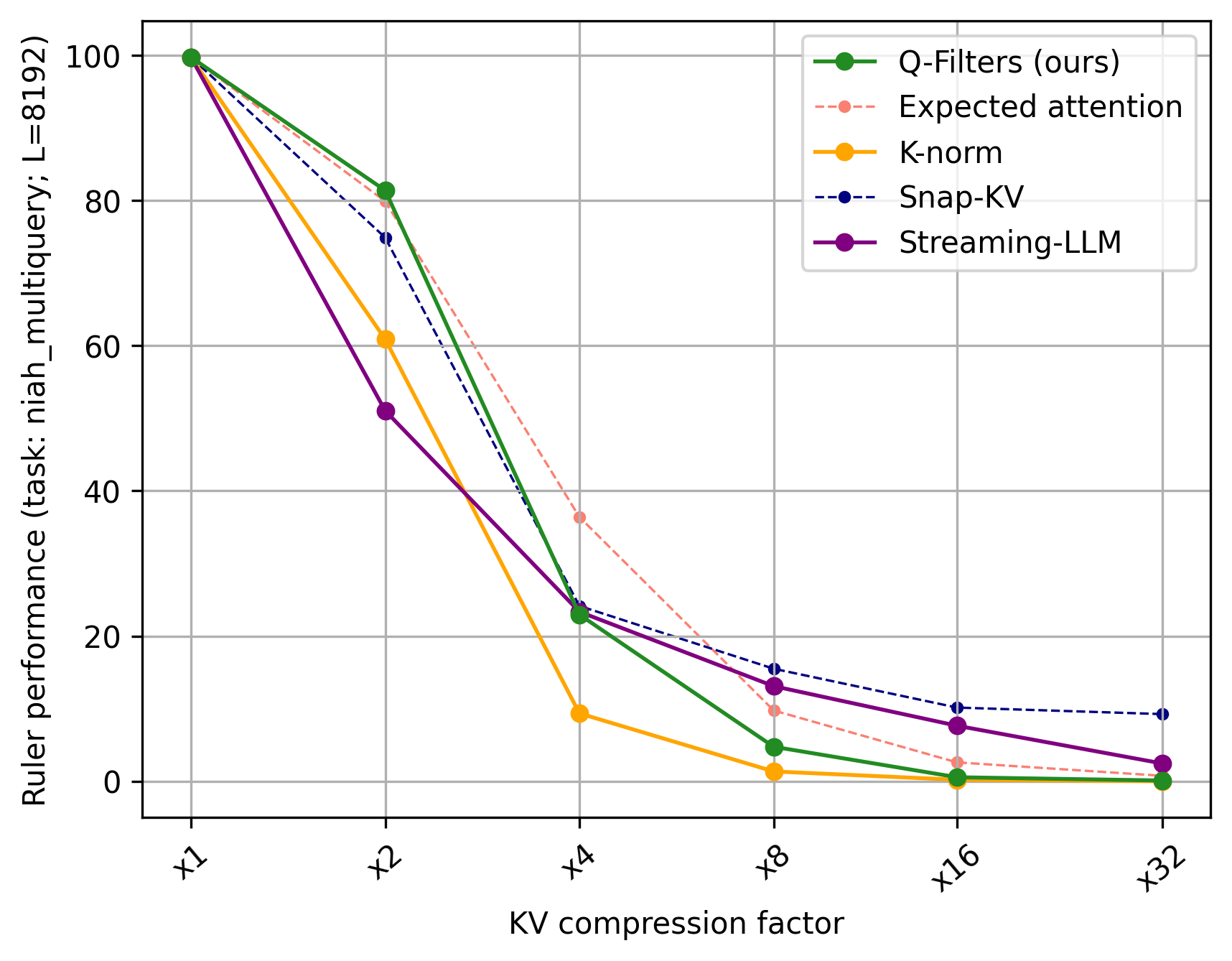}
         \caption{NIAH - Multi-Query}
         \label{fig:ruler_multiquery}
     \end{subfigure}
     \begin{subfigure}[h]{0.32\textwidth}
         \centering
         \includegraphics[width=\textwidth]{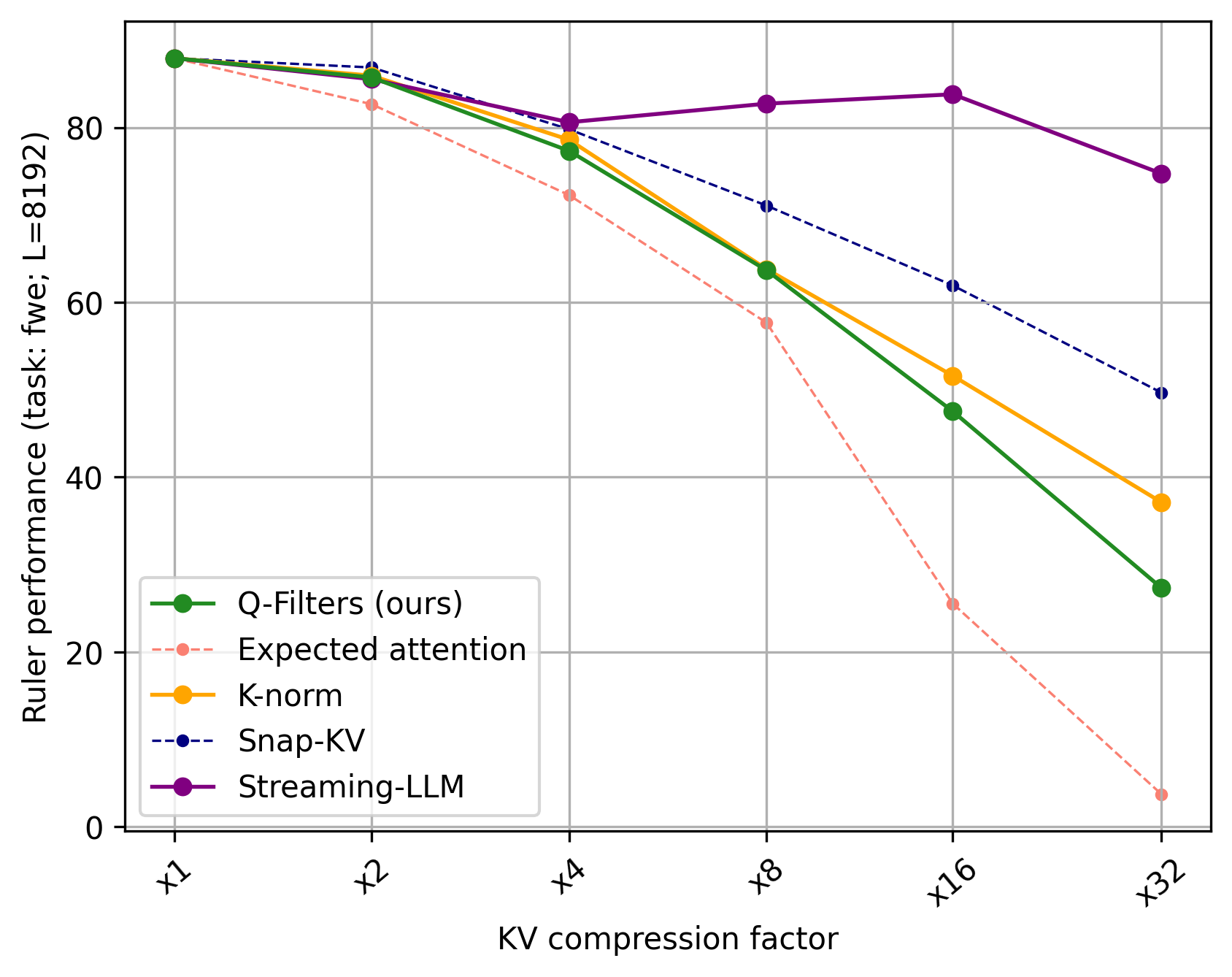}
         \caption{Frequent Words Extraction (FWE)}
         \label{fig:ruler_fwe}
     \end{subfigure}
     \begin{subfigure}[h]{0.32\textwidth}
         \centering
         \includegraphics[width=\textwidth]{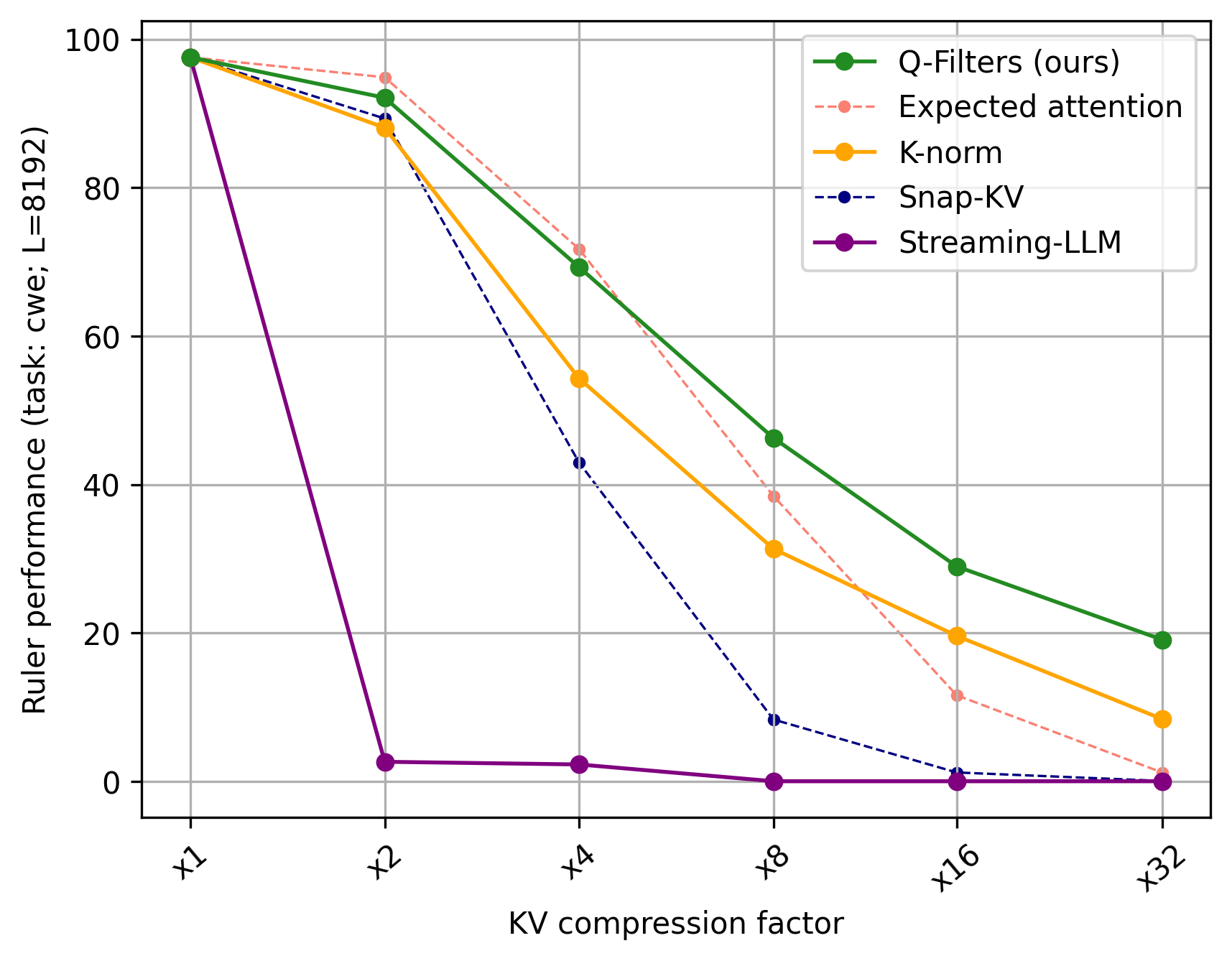}
         \caption{Common Words Extraction (CWE)}
         \label{fig:ruler_fwe}
     \end{subfigure}
        \caption{Performance of Llama-3.1-8B-Instruct using several \kvcache compression methods on individual tasks from the Ruler dataset (with length 8192) as compression ratio evolves. We report prompt compression methods using dotted lines for comparison.}
        \label{fig:ruler_details}
\end{figure*}

In \Cref{fig:ruler_details} we report detailed evaluation on the subsets of the Ruler dataset \citet{hsieh2024ruler}.

\section{Generation examples}
Using Llama-3.1-8B, we identify interesting cases where \qfilters provide the correct next token in a given long context, while K-norm and Streaming-LLM fail to capture the relevant information.

\begin{table*}[h!]
\centering
\small
\begin{tabular}{|>{\arraybackslash}m{0.4\textwidth}|>{\arraybackslash}m{0.18\textwidth}|>{\arraybackslash}m{0.18\textwidth}|>{\arraybackslash}m{0.18\textwidth}|}
\hline
\textbf{Text Sample (Context)} & \textbf{Q-Filters} & \textbf{K-Norm} & \textbf{Streaming-LLM} \\
\hline
\textit{One of the show's first longest-running storylines was the rivalry between a young manicurist Jill Foster Abbott (Brenda Dickson, Jess Walton) and wealthy socialite, \textbf{Katherine} Chancellor (\textbf{Jeanne} Cooper). [...] After much investigation, it is revealed that Kay is Jill's biological...} & \textcolor{Green}{\textit{mother}}  & \textcolor{red}{\textit{father}} & \textcolor{red}{\textit{father}} \\
\hline
\textit{Both extreme right-wing leaders taught and practised the theology of \textbf{Christian Identity}, a belief system which the FBI includes on its watch list as an extremist religion. [...] Here, the group trained an estimated 1,500 of like-minded Christian...} & \textcolor{Green}{\textit{Identity}} & \textcolor{red}{\textit{fundamental}} & \textcolor{red}{\textit{fundamental}} \\
\hline
\textit{The Viral Fever\newline [...] \textbf{TVF} debuted their platform, releasing the final two episodes of Pitchers on \textbf{TVFPlay}. [...] \textbf{TVF} claims to have worked with over 150 brands. [...] The show has been on hold as writer Biswapati Sarkar focuses on writing web series, including the sequel to TV...} & \textcolor{Green}{\textit{F}} & \textcolor{red}{\textit{\_show}} & \textcolor{red}{\textit{\_show}} \\
\hline
\end{tabular}
\caption{Next-token generation examples for different KV Cache Compression methods, applied to Wikipedia article. Passages in bold correspond to useful information that is necessary to resolve the ambiguity in the choice of the next token.}
\label{tab:kv-cache-examples}
\end{table*}

\section{Implementation Details}
For all our experiments, we use the popular Huggingface models with the recently released KVPress library \cite{kvpress2024}. 


\end{document}